\pdfoutput=1
\documentclass[11pt]{article}

\usepackage{caption}
\usepackage{graphicx}
\usepackage{subfigure}
\usepackage{multirow}
\usepackage{url}
\usepackage{acl}

\usepackage{times}
\usepackage{latexsym}
\usepackage{amsthm,amsmath,amssymb}
\usepackage{mathrsfs}
\usepackage{booktabs}
\usepackage[normalem]{ulem}

\usepackage[T1]{fontenc}
\usepackage[utf8]{inputenc}

\usepackage{microtype}
\usepackage{enumitem}
\usepackage{color}
\usepackage{soul}
\usepackage{makecell}
\usepackage{adjustbox}
\usepackage{multirow}
\usepackage{amsmath}
\usepackage{amssymb}

\usepackage{xcolor}
\newcommand{\greyhighlight}[1]{\colorbox{gray!20}{#1}}

\usepackage{easyReview}
\usepackage{inconsolata}
\makeatletter
\def\@fnsymbol#1{\ensuremath{\ifcase#1\or \dagger\or \ddagger\or
   \mathsection\or \mathparagraph\or \|\or **\or \dagger\dagger
   \or \ddagger\ddagger \else\@ctrerr\fi}}
    \makeatother

\title{CLAMBER: A Benchmark of Identifying and Clarifying Ambiguous Information Needs in Large Language Models}

\author{
Tong Zhang$^{\spadesuit\heartsuit\star}$, \quad
Peixin Qin$^{\spadesuit\heartsuit\star}$
, \quad
Yang Deng$^{\clubsuit}$, \quad 
\textbf{Chen Huang}$^{\spadesuit\heartsuit}$, \quad \\
\textbf{Hongru Liang}$^{\spadesuit\heartsuit}$, \quad
\textbf{Junhong Liu}$^{\diamondsuit}$, \quad 
\textbf{Dingnan Jin}$^{\diamondsuit}$, \quad \\
\textbf{Wenqiang Lei}$^{\spadesuit\heartsuit}$\thanks{Corresponding author.}, \quad
\textbf{Tat-Seng Chua}$^{\clubsuit}$
\\
${\spadesuit}$ College of Computer Science, Sichuan University, China \\
${\heartsuit}$ Engineering Research Center of Machine Learning and Industry Intelligence, \\Ministry of Education, China \\
\quad ${\diamondsuit}$ Ant Group \quad ${\clubsuit}$ National University of Singapore \\
\{scu.zhangtong, huangc.scu\}@gmail.com \quad wenqianglei@scu.edu.cn\\ 
}

\begin{document}
\maketitle
\def\thefootnote{$\star$}\footnotetext{Both authors contributed equally to this study.}\def\thefootnote{\arabic{footnote}}

\begin{abstract}
Large language models (LLMs) are increasingly used to meet user information needs, but their effectiveness in dealing with user queries that contain various types of ambiguity remains unknown, ultimately risking user trust and satisfaction. 
To this end, we introduce CLAMBER, a benchmark for evaluating LLMs using a well-organized taxonomy. 
Building upon the taxonomy, we construct $\sim 12K$ high-quality data to assess the strengths, weaknesses, and potential risks of various off-the-shelf LLMs.
Our findings indicate the limited practical utility of current LLMs in identifying and clarifying ambiguous user queries, even enhanced by chain-of-thought (CoT) and few-shot prompting. 
These techniques may result in overconfidence in LLMs and yield only marginal enhancements in identifying ambiguity. 
Furthermore, current LLMs fall short in generating high-quality clarifying questions due to a lack of conflict resolution and inaccurate utilization of inherent knowledge.
In this paper, CLAMBER presents a guidance and promotes further research on proactive and trustworthy LLMs. Our dataset is available at \url{https://github.com/zt991211/CLAMBER}.
\end{abstract}

\section{Introduction}

Given well-defined user queries, large language models (LLMs) have demonstrated remarkable proficiency in facilitating the information search process \citep{pan2023kwaiagents, kamalloo2023hagrid, zhang2023clarify, huang-etal-2023-reduce}.
They provide more precise search results with the help of the inherent knowledge stored within LLMs.
Nonetheless, as evidenced by previous studies \citep{kuhn2023clam, deng2023rethinking}, the practical utility of LLMs is hindered by unclear and ambiguous user queries in real-world scenarios. 
For instance, in a query like "\textit{what are the strategies for saving?}", the term "\textit{saving}" can have multiple interpretations, such as "\textit{saving money}" or "\textit{saving from sins}", depending on the user's actual need.
This necessitates LLMs proactively identifying (i.e., determine if the query is ambiguous or not) and clarifying the ambiguities rather than providing potentially incorrect answers that may not align with the user's true needs, ultimately risking user trust and satisfaction \citep{liao2023proactive}. 

%As a notable example, they are confined to pre-LLM era perspectives \dy{(what is pre-LLM era perspectives?)} and overlook the LLM-oriented ambiguity that may occur when inherent knowledge stored within LLMs have conflict understanding about the query.
Driven by this concern, recent works have explored LLMs' capacity to address ambiguous queries \citep{deng-etal-2023-prompting, kuhn2023clam}. 
However, these investigations have been somewhat fragmented, lacking a comprehensive taxonomy, leading to incomplete and inconsistent handling of ambiguity distributions \citep{keyvan2022approach, rahmani-etal-2023-survey}. 
As a notable example, they are often limited to contextual ambiguity, where the given context is insufficient for producing a definitive answer. 
In the era of LLMs, there should be more emphasis on the LLM-oriented ambiguity that may occur when inherent knowledge stored within LLMs have conflict understanding about the query.
Consequently, it still remains unclear which ambiguities LLMs can effectively identify and clarify, along with the challenges that LLMs persistently encounter in this regard.

To this end, we introduce CLAMBER (\textbf{Cl}arifying \textbf{Amb}iguous Qu\textbf{er}y), a novel benchmark for comprehensively evaluating LLMs in identifying and clarifying various ambiguities using a well-organized taxonomy.
Drawing inspiration from the input-process-output framework for evaluating collaborative systems \cite{pinsonneault1989impact}, we establish a taxonomy that consolidates both input understanding and task completion perspectives into three primary dimensions, as illustrated in Table \ref{CLAMBER_Taxonomy}. 
These dimensions are further conceptualized into eight fine-grained categories to facilitate in-depth evaluation. 
Building upon this taxonomy, we construct $\sim 12K$ data for analyzing the pros and cons of LLMs when identifying and clarifying ambiguities.
%CLAMBER includes GPT-4-generated and human-verified $\sim 12K$ data, which are used to analyze the pros and cons of LLMs when clarifying different ambiguities. \dy{better not to mention GPT-generated data too much.}

With CLAMBER, we comprehensively evaluate strengths, weaknesses, and potential risks of various LLMs.
Our findings indicate that ChatGPT \citep{ChatGPT} outperforms other small-scale LLMs, especially excelling in identifying and clarifying ambiguities in multifaceted queries \citep{clarke2009overview}, such as "\textit{What is the largest manufacturer in China?}", which does not specify the type of "\textit{manufacturer}". 
However, they still encounter numerous challenges: 1) \textbf{current LLMs, despite leveraging chain-of-thought (CoT) and few-shot prompting, face challenges in identifying ambiguities.}
Our results suggest that CoT and few-shot prompting may lead to the over-confidence issue in small-scale LLMs, impacting ambiguity identification negatively.
Even with a large number of shots and CoT support, LLMs only achieve a marginal improvement.
Moreover, current LLMs struggle to leverage contextual cues to disambiguate pronouns, highlighting the inadequacy in deducing underlying ambiguities.
%Furthermore, even when bolstered by a large number of shots and CoT, the slight enhancement achieved through few-shot prompting is minimal. 
%\dy{Moreover, there is only minimal improvement even with a large number of demonstration samples for few-shot prompting.} 
%is challenging to teach LLMs to fully comprehend the correct ambiguity reasoning. Providing large diverse examples and integrating them with CoT can somewhat attain a weak enhancement.
2) \textbf{Current LLMs fail to ask high-quality clarifying questions, due to the inability of knowing their knowledge gap.} 
Despite LLMs recognize a query containing ambiguities, their lack of conflict resolution and inaccurate use of inherent knowledge results in uncertainty about which ambiguity to clarify.
This prompts the need of developing effective methods for LLMs to resolve conflicts and accurately utilize their inherent knowledge.

In this paper, CLAMBER stands as a valuable resource to provide guidance and insight into evaluating LLMs and addressing ambiguous information needs for future improvements. In conclusion, our contributions are threefold: 
\vspace{-\topsep}
\begin{itemize}[leftmargin=*, itemindent=0.05cm, itemsep=-4pt]
    \item We introduce a taxonomy for categorizing various query ambiguities. This taxonomy combines three primary dimensions, detailed as eight categories for facilitating fine-grained evaluations.
    \item We present a novel benchmark called CLAMBER, tailored to the characteristics of LLMs. It contains $\sim 12K$ data featuring ambiguous user queries across diverse categories.
    \item With CLAMBER, we evaluate the off-the-shelf LLMs in an inclusive manner. Our findings shed light on why current LLMs struggle to identify and clarify ambiguities. These insights will guide future research in this field.
\end{itemize}

\begin{table*}[ht]
\centering
\resizebox{\linewidth}{!}{
\begin{tabular}{c|l|l|l}
\hline
\multicolumn{1}{l|}{Dimension}                                            & Category      & Explanation                                                                                              & Example                                                                                                                                                                                                                                  \\ \hline
\multirow{2}{*}{\begin{tabular}[c]{@{}c@{}} \\ Epistemic \\ Misalignment\end{tabular}} & \textbf{UNFAMILIAR}    & \begin{tabular}[c]{@{}l@{}}Query contains unfamiliar \\ entities or facts\end{tabular}                            & Find the price of Samsung Chromecast.                                                                                                                                                                                                             \\ \cline{2-4} 
                                                                                   & \textbf{CONTRADICTION} & \begin{tabular}[c]{@{}l@{}}Query contains self- \\ contradictions\end{tabular}                                     & \begin{tabular}[c]{@{}l@{}}Output 'X' if the sentence contains [category withhold] and 'Y' otherwise. \\ The critic is in the restaurant.\textgreater{}X. The butterfly is in the river.\textgreater{}Y. \\ The boar is in the theatre.\textgreater{}?\end{tabular} \\ \hline
\multirow{3}{*}{\begin{tabular}[c]{@{}c@{}} Linguistic \\ Ambiguity\end{tabular}}   & \textbf{LEXICAL}       & \begin{tabular}[c]{@{}l@{}}Query contains terms \\ with multiple meanings\end{tabular}                            & Tell me about the source of Nile.                                                                                                                                                                                                                 \\ \cline{2-4} 
                                                                                   & \textbf{SEMANTIC}      & \begin{tabular}[c]{@{}l@{}}Query lacks of context\\ leading multiple interpretations\end{tabular}                 & When did he land on the moon?                                                                                                                                                                                                                     \\ \hline
\multirow{7}{*}{\begin{tabular}[c]{@{}c@{}} \\ \\ \\ Aleatoric \\ Output\end{tabular}}       & \textbf{WHO}           & \begin{tabular}[c]{@{}l@{}}Query output contains \\ confusion due to \\ missing personal elements\end{tabular}    & Suggest me some gifts for my mother.                                                                                                                                                                                                              \\ \cline{2-4} 
                                                                                   & \textbf{WHEN}          & \begin{tabular}[c]{@{}l@{}}Query output contains\\ confusion due to\\ missing temporal elements\end{tabular}      & How many goals did Argentina score in the World Cup?                                                                                                                                                                                                        \\ \cline{2-4} 
                                                                                   & \textbf{WHERE}         & \begin{tabular}[c]{@{}l@{}}Query output contains\\ confusion due to\\ missing spatial elements\end{tabular}       & Tell me how to reach New York.                                                                                                                                                                                                                    \\ \cline{2-4} 
                                                                                   & \textbf{WHAT}          & \begin{tabular}[c]{@{}l@{}}Query output contains\\ confusion due to\\ missing task-specific elements\end{tabular} & Real name of gwen stacy in spiderman?                                                                                                                                                                                                             \\ \hline
\end{tabular}
}
\caption{The proposed taxonomy of ambiguous queries and examples. The clarifying questions of each example are provided in Table \ref{tab:example_cq}.}
\label{CLAMBER_Taxonomy}
% \vspace{-4mm}
\end{table*}

\section{Related Works}
Our research is closely tied to the taxonomy and resolution of ambiguities in LLMs. We provide a literature review and highlight our differences. 

% Our objective is to investigate the capacity of LLMs in identifying and clarifying diverse ambiguous queries. Consequently, we categorize the related works into two sections: 1) the first section discusses ambiguity in information retrieval, outlining the ambiguities present in previous fragmented works and shedding light on new ambiguities, particularly for LLMs. 2) the section section concentrates on recent approaches aimed at enhancing LLMs' ability to identify ambiguous queries and formulate clarifying inquiries.

% Ambiguity is a fundamental feature of language \citep{wasow2005puzzle}, allowing speakers to balance efficiency and clarity in communication \citep{zipf2016human, piantadosi2012communicative}.
% These tasks include question answering \citep{kumar-black-2020-clarq, min-etal-2020-ambigqa}, search \citep{song2007identifying, zamani2020mimics, aliannejadi2021building} and recommendation \citep{Ren_2021}.

% Ambigutiy is a longstanding and widely-studied issue in information retrieval \citep{keyvan2022approach, rahmani-etal-2023-survey}.
% As a result, numerous datasets are introduced, such as clariq \citep{aliannejadi2021building}, ambigqa \citep{min-etal-2020-ambigqa} and mimics \citep{zamani2020mimics}.
% However, two obstacles prevent these datasets from serving as a standardized benchmark for comprehensively evaluating LLMs to address various ambiguities:

\noindent \textbf{Ambiguity Taxonomy.} As evidenced by a recent survey \citep{rahmani-etal-2023-survey}, there is a lack of a well-organized taxonomy for ambiguity in information retrieval. 
While previous research attempts to integrate ambiguity taxonomies, their taxonomies are fragmented and underdeveloped \citep{ginzburg1996interrogatives, song2007identifying}, failing to facilitate comprehensive evaluations. 
Recent taxonomies \citet{min-etal-2020-ambigqa, guo2021abg} are formulated based on a limited set of factual questions and lack precise definitions for each category. 
Moreover, existing taxonomies were established before the era of LLMs, disregarding the ambiguity specific to LLMs that may arise from conflicting interpretations of queries by the inherent knowledge stored within LLMs.
This is evident when LLMs encounter unfamiliar entities \citep{yin2023alcuna} or potential inconsistencies within queries \citep{tamkin2022task}. 
For the first time, we introduce a well-organized taxonomy for categorizing various query ambiguities. Our taxonomy draws inspiration from the input-process-output view to evaluate collaborative systems. It combines three primary dimensions that capture potential ambiguities during input understanding and task completion of LLMs. 
Using this taxonomy, we construct $\sim 12K$ data for analyzing the pros and cons of LLMs in resolving different ambiguities.
\noindent \textbf{Resolving ambiguity in LLMs.}
Recent efforts resort to CoT and few-shot prompting to enhance LLMs' capacity in identifying and clarifying ambiguous queries \citep{deng-etal-2023-prompting, kuhn2023clam, cole-etal-2023-selectively}. 
While these efforts have shown some improvements in performance, they are confined to tasks involving specific types of ambiguities, such as lexical ambiguity. 
In this paper, we incorporate CoT and few-shot prompting as baselines to evaluate their efficacy and inadequacy across a broader range of ambiguity types using CLAMBER.
Other related works try to examine which of the two queries exhibits more ambiguity \citep{zhang2023clarify}, unable to determine if a query is ambiguous.

% In terms of identifying ambiguous queries, \citep{trienes2019identifying} utilizes logistic regression to identify ambiguous queries based on similar query characteristics, while \citep{dhole2020resolving} uses a BiLSTM model to differentiate between ambiguous and unambiguous queries. 
% In terms of generating clarifying questions, both rule-based and neural network-based methods have been suggested. For example, \citep{wang2021template, wang2023zeroshot} utilize a template for clarifying questions that is completed with words from a vocabulary. 
% \citep{shridhar-etal-2023-distilling} investigates the generation of clarifying questions as a supervised learning task, using the questions for multi-step reasoning in knowledge distillation, while \citep{pyatkin-etal-2023-clarifydelphi} uses reinforcement learning (RL) to guide their clarifying question generation.

% \citep{dhole2020resolving} approaches disambiguation by distinguishing between different plausible user intents for a given question through syntactic transformations, while \citep{rao-daume-iii-2019-answer} use a GAN to generate clarifying questions.

\section{CLAMBER Benchmark}

To evaluate LLMs in an inclusive manner, we present CLAMBER, which introduces a taxonomy encompassing three key dimensions (i.e., \textit{Epistemic Misalignment}, \textit{Linguistic Ambiguity}, \textit{Aleatoric Output}) that capture potential ambiguities during input understanding and task completion. 
These three dimensions are further divided into eight specific categories. 
We delve into the taxonomy and data collection process in following sections. 
Each data comprises a user query, a binary ambiguity label, and a clarifying question for ambiguous queries.
See details of data collection in Appendix \ref{detail_data_collection}.

% Our taxonomy covers three dimensions of ambiguity: Epistemic Misalignment (EM), Linguistic Ambiguity (LA), and Aleatoric Output (AO). 
% The first two dimensions of ambiguity arise from challenges in comprehending the input.
% The third dimension of ambiguity occurs when the input is well-formed but lacks certain elements required for generating results.

% 要写generated by GPT-4

\subsection{Epistemic Misalignment (EM)}
Building upon the input understanding perspective, EM occurs when inherent knowledge stored within LLMs have conflict understanding about the query \citep{cole-etal-2023-selectively, zhang2023clarify}.
This ambiguity is a distinctive feature of LLMs, as they respond to queries using their inherent knowledge. We categorize EM into two categories based on the source of conflicting:
\vspace{-\topsep}
\begin{itemize}[leftmargin=*, itemindent=0.05cm, itemsep=-4pt]
    \item \textbf{Unfamiliar}. It refers to situations where LLMs encounter entities or facts that are unfamiliar to them, either because they are not within the LLMs' inherent knowledge or because they contradict it. 
    Given a query "\textit{Find the price of Samsung Chromecast}", if LLMs only have inherent knowledge on "\textit{Google Chromecast}" or "\textit{Samsung Chromebook}" and are unfamiliar with "\textit{Samsung Chromecast}", LLMs should proactively ask for clarification about "\textit{Samsung Chromecast}" rather than provide answers regarding "\textit{Google Chromecast}" or "\textit{Samsung Chromebook}", ultimately risking user satisfaction.
    \item \textbf{Contradiction}. It refers to situations where LLMs infers contradictions within queries based on their inherent knowledge.
    For example, given a query "\textit{Output 'X' if the sentence contains [category withhold] and 'Y' otherwise. Examples: The critic is in the restaurant.>X. The butterfly is in the river.>Y. Sentence: The boar is in the theatre.>?}", LLMs may infer two different categories (i.e., human and indoor location) from provided examples. 
    This contradiction could lead to confusion for LLMs. Consequently, LLMs should seek clarification by asking: "\textit{Does this category a human or an indoor location?}"
\end{itemize}

\noindent \textbf{Data Collection}. 
To evaluate the \textit{Unfamiliar} category, it is important to determine exactly what LLMs are unfamiliar with \citep{wang2023resolving}.
To mitigate bias stemming by training data, CLAMBER opts to utilize entirely new, fabricated knowledge that are unfamiliar to all LLMs. 
To achieve this, we resort to the ALCUNA dataset \citep{yin2023alcuna} as our data resource, which includes queries that contain new entities fabricated by modifying existing ones. 
We classify the queries containing new entities as ambiguous, while the rest are unambiguous. 
Subsequently, we instruct GPT-4 to generate a clarifying question for each ambiguous query, focusing on the ambiguity of new entities.
As for the \textit{Contradiction} category, the contradiction in CLAMBER occurs when the query and the given examples fail to match within a single interpretation.
To achieve this, we directly utilize the AmbiTask dataset \citep{tamkin2022task} to provide ambiguous queries, which encodes contradiction among queries and provided examples. Additionally, we create clarifying questions for ambiguous queries by rule-based templates and manually transform ambiguous queries into unambiguous ones by resolving contradictions.

% For Internal Unfamiliar, we use the ALCUNA Dataset \citep{yin2023alcuna} to analyze queries that contain new entities created by modifying existing ones. \highlight{I don't know what ALCUNA is? why you use it. and i dont understand the meaning of "analyze queries that contain new entities created by modifying existing ones"}
% We classify queries with new entities as ambiguous, while those with existing entities as unambiguous.
% Subsequently, we generate a clarifying question for each ambiguous query.
% For External Inconsistent, we employ the AmbiTask Dataset \citep{tamkin2022task}. \highlight{I don't know what AmbiTask is? why you use it.}
% This dataset creates multiple classification tasks based on queries and a few labeled examples. 
% Ambiguity arises when queries and labeled examples align with more than one task, leading to contradictions. 
% We directly incorporate ambiguous queries and transform them into unambiguous queries by resolving contradictions
% Additionally, we create clarifying questions for ambiguous queries.

\subsection{Linguistic Ambiguity (LA)}

% Here, we focus on two types of linguistic ambiguity, namely lexical and semantic, as they encapsulate the main challenges faced by modern CIS systems \citep{xu2019asking, guo2021abg}.
% The lexical level concerns individual words with multiple meanings (e.g., Homonymy and Polysemy), while the semantic level involves the absence of context leading to more than one interpretation of a sentence \citep{amna2022ambiguity}.
% There are also more situations that present linguistic ambiguity, such as pragmatics and phonetic errors \citep{How2013Jiang}, but these are currently far from being resolved.

Building upon the input understanding perspective, LA arises when a word, phrase, or statement can be interpreted in multiple ways due to its imprecise or unclear meaning \citep{berry2004ambiguity, ortega2023linguistic}. 
We categorize LA into the the lexical and semantic ambiguities\footnote{We omit the syntactic and pragmatic ambiguities as they are not commonly used in information retrieval.}, which encapsulate the main challenges in information retrieval \citep{coden2015did, xu2019asking}.  
\vspace{-\topsep}
\begin{itemize}[leftmargin=*, itemindent=0.05cm, itemsep=-4pt]
    \item \textbf{Lexical Ambiguity}. It concerns individual terms with multiple meanings. 
    For example, given a query "\textit{Tell me about the source of Nile}", the term "\textit{source of Nile}" can be interpreted in two meanings: the origin of the Nile river or the board game named "\textit{source of Nile}". In this case, LLMs should ask for clarification: "\textit{Are you referring to the Nile river or the board game?}"
    \item \textbf{Semantic Ambiguity}. It involves the lack of context leading to more than one interpretation of a sentence \citep{ortega2023linguistic}. 
    For example, given a query "\textit{When did he land on the moon?}", it is unclear who "\textit{he}" may refer to without context. In this case, LLMs should ask for clarification: "\textit{Who is 'he' referring to?}" 
\end{itemize}

\noindent \textbf{Data Collection}. 
Lexical Ambiguity pertains to individual terms with multiple meanings, often found in entity names and polysemy words \citep{keyvan2022approach}. In this paper, we resort to the AmbER \citep{chen-etal-2021-evaluating} and AmbiPun dataset \citep{mittal2022ambipun}, which contain ambiguous entity names and ambiguous polysemy words, respectively. 
We extract these terms along with their various meanings from the datasets and then create ambiguous queries, clarifying questions and unambiguous queries using GPT-4. 
As for Semantic Ambiguity, CLAMBER pay special focus investigating referent ambiguity following \citep{kuhn2022clam, ortega2023linguistic}. This type of ambiguity occurs in queries containing pronouns that lack contextual clues for clarification. Specifically, we employ the AmbiCoref dataset \citep{yuan2023ambicoref}, which consists of minimal pairs featuring ambiguous and unambiguous referents.
In this regard, an ambiguous query can be achieved by reducing context sizes to a single sentence and creating sentences where the verbs involved limit the interpretation of their arguments. 
Additionally, we obtain the unambiguous queries by instructing GPT-4 and obtain clarifying questions by rule-based templates.

\subsection{Aleatoric Output (AO)}
Building upon the task completion perspective, AO occurs when the input is well-formed but the output contains potential confusion due to the lack of essential elements.
It is prevalent across various types of queries in information retrieval, including faceted queries \citep{clarke2009overview}, queries missing details \citep{trienes2019identifying}, board queries \citep{song2007identifying} and under-specific queries \citep{aliannejadi2021building}.
Previous studies have focused on specific aspects of this ambiguity, but there is a need for a more comprehensive understanding of this ambiguity in order to advance research.
Inspired by \citep{zamani2020generating}, we categorize AO into four specific categories based on the type of missing elements: 
\vspace{-\topsep}
\begin{itemize}[leftmargin=*, itemindent=0.05cm, itemsep=-4pt]
    \item \textbf{Whom} denotes the absence of personal details, such as expertise. Given a query "\textit{Suggest me some gifts for my mother}", the response may vary due to missing the personal preferences of his mother. In this case, a clarifying question like: "\textit{What specific preferences does your mother have?}" would be preferred.
    \item \textbf{Where} pertains to the lack of spatial information, such as departure place. For example, given a query "\textit{Tell me how to reach New York}", the response may vary due to missing the specific departure information. In this case, LLMs should ask for clarification "\textit{Where do you start from?}"
    \item \textbf{When} refers to the absence of temporal elements, such as specific dates. Given a query "\textit{How many goals did Argentina score in the World Cup?}", the response may vary due to missing the specific World Cup year. This ambiguity requires LLMs to seek further details by asking clarifying questions "\textit{Which year of the World Cup are you referring to?}"
    \item \textbf{What} refers to the remaining types. For example, when a query is "\textit{Who played Thanos in Guardians of the Galaxy?}", the response may vary due to missing the specific version of Guardians of the Galaxy. Clarifying question should arise: "\textit{Which version are you referring to: TV series, 2014 film, or Telltale Series?}"
    % \highlight{Example and why} are collectively referred to as "What", encompassing task-specific elements.
\end{itemize}

\noindent \textbf{Data Collection}.
We construct four categories of ambiguities by recognizing the specific missing elements in well-structured queries. 
To accomplish this, we resort to the the AmbigQA dataset \citep{min-etal-2020-ambigqa} and the Dolly-16K dataset \citep{DatabricksBlog2023DollyV2}) containing factual and instrumental user search intent \citep{ORCAS-I2022Alexander}.
As for AmbigQA dataset, queries with multiple answers are deemed ambiguous, while those with a single answer are considered unambiguous. Ambiguous queries are manually categorized into the four categories. 
In the Dolly dataset, each query is automatically labeled as ambiguous or unambiguous by GPT-4, then manually verified and classified into the four categories if marked as ambiguous. Due to the difficulty in crafting category-specific unambiguous queries, all four categories share the same set of unambiguous queries.
% Then, we follow the clarifying annotations in \citep{lee2023asking} and manually categorize the ambiguous queries with annotated clarifying questions into our four categories.
%we instruct GPT-4 to automatically classify these queries into ambiguous and unambiguous, and then generate clarifying questions for ambiguous queries.
%Similar to AmbigQA, we further manually categorize the ambiguous queries into our four categories.

\subsection{Validation and Revision}
To ensure the quality of our dataset, we engage five linguistic experts for validation and revision. 
Initially, each data is validated by four experts, and subsequently consolidated by the remaining expert. 
The validation process includes verifying the accuracy of ambiguity labels and assessing the effectiveness of clarifying questions. 
If there are discrepancies between the four experts' validation, the final expert examines their feedback and implements necessary data revisions.
For further details on the validation and revision procedures, please refer to Appendix \ref{Manual_Revision}.
Finally, our data statistics are presented in Table \ref{data_statistics}.

% \begin{table}[ht]
% \centering
% \caption{The dataset statistics.}
% \label{data_statistics}
% \resizebox{\linewidth}{!}{
% \begin{tabular}{c|c|c|c}
% \hline
% \textbf{Category} & \textbf{Ambiguous} & \textbf{Unambiguous}  & \multicolumn{1}{l}{\textbf{Total}} \\ \hline
% Unfamiliar        & 600                & 600                   & 1200                               \\ \hline
% Contradiction     & 600                & 600                   & 1200                               \\ \hline
% Lexical           & 900                & 900                   & 1800                               \\ \hline
% Semantic          & 400                & 400                   & 800                                \\ \hline
% What              & 1255               & \multirow{4}{*}{3884} & \multirow{4}{*}{7167}              \\ \cline{1-2}
% Whom              & 762                &                       &                                    \\ \cline{1-2}
% When              & 779                &                       &                                    \\ \cline{1-2}
% Where             & 487                &                       &                                    \\ \hline
% \end{tabular}
% }
% \end{table}

\begin{table}[t]
  \centering
   \scalebox{0.6}{
  \begin{tabular}{ccccc}
    \toprule
    \multirow{2}{*}{Category}  &  \multirow{2}{*}{Sources}& \multicolumn{3}{c}{Distribution} \\
    \cmidrule(r){3-5}
    &&Ambig. & Non-Ambig. & ALL \\
    \midrule
    Unfamiliar & ALCUNA &684 &547 &1231 \\
    Contradiction & AmbiTask &600 &600 &1200  \\ 
    Lexical & AmbER,AmbiPun &815  &921 &1,736 \\
    Semantic & AmbiCoref &400 &400 &800  \\
    What & AmbigQA, Dolly &1255  &\multirow{4}{*}{3884 in total} & \multirow{4}{*}{7167 in total} \\
    Whom & AmbigQA, Dolly & 762 & & \\
    When & AmbigQA, Dolly & 779 & & \\
    Where & AmbigQA, Dolly & 487 & & \\
    % All&& &8,713 &9,619 &3,608 &21,940\\
    \bottomrule
\end{tabular}}
\caption{CLAMBER Dataset Sources and Statistics.}
\label{data_statistics}
% \vspace{-6mm}
\end{table}

\section{Experimental Design}
% \vspace{-1mm}
We consider two tasks to evaluate off-the-shelf LLMs, including identifying ambiguities (cf. Section \ref{task1}) and asking clarifying questions (cf. Section \ref{task2}). Each task utilizes different evaluation metrics, outlined in the corresponding sections.

% Given a query Q, LLMs first aim to predict whether clarification is required by classifying Q as ambiguous or unambiguous.
% If ambiguity is identified, LLMs should then generate an clarifying question to resolve the ambiguity.

% 具体数据集的数目，然后介绍两部分的实验细节

% \subsection{Experimental Setups}

\noindent \textbf{Test Dataset.}
Our experiments are conducted on sub-sample of 3600 instances randomly selected, preserving the same number of data samples per category.
There are 200 positive and negative examples for each category.
Particularly, the negative examples of each category within Aleatoric Output is 800 since its uniform nature.

%\noindent \textbf{Evaluation Metrics.} 
% To evaluate identifying ambiguity, following \citep{hu2023large, deng-etal-2023-prompting}, we adopt the Accuracy and F1 score as metrics. Specifically, we opt the weighted F1 due to the balanced nature of our test dataset. To evaluate clarifying questions, we apply BertScore \citep{zhang2019bertscore} for automatic evaluation, as lexical matching metrics may fail to fully capture clarification capabilities \citep{guo2021abg}. We also conduct human evaluation to score the utility of generated questions in resolving ambiguities (\textbf{Help.}).

% Our goal is to maximize model accuracy in identifying ambiguity while minimizing the need for excessive clarification.
% Our aim is to generate clarifying questions that are natural and helpful for eliminating ambiguity.

% For identifying ambiguities, following \citep{hu2023large, deng-etal-2023-prompting}, we adopt the Accuracy and F1 scores as the evaluation metrics of identifying ambiguities.
% Specifically, we use the weighted F1 score as our metric, given the balanced nature of our test set.
% Our aim is to ensure the model's accuracy without ambiguity, minimizing the need for excessive clarification.
% For asking clarifying questions, we adopt the BertScore as the metric since the automatic lexical matching metrics may fail to actually estimate the clarification capability of the generated clarifying questions \citep{guo2021abg}, we also adopt human evaluation to score whether the generated question is useful for
% clarifying the existing ambiguity (Use.).

\noindent \textbf{Usage of LLMs.} 
As our set of LLMs, we evaluate Llama2-13B-Chat (i.e., Llama2-13B), Llama2-13B-Instruct (i.e., Llama2-13B-I), Vicuna-13B, Llama2-70B-Chat (i.e., Llama2-70B), and the GPT-3.5-Turbo-16k (i.e., ChatGPT). 
These LLMs are widely used in recent studies of information search \cite{deng-etal-2023-prompting, zhang2023clarify}.
% To facilitate reproducibility, we set the temperature to 0 for generating the deterministic outputs with the same inputs.
% In addition, we set the maximum number of new tokens to 128.

\noindent \textbf{Prompting Schemes.}
Following \citep{deng-etal-2023-prompting}, we devise four prompting schemes for evaluation: 
1) Zero-shot w/o CoT, where the LLM is evaluated directly on the test dataset, 
2) Zero-shot w/ CoT \citep{wei2022chain}, where the LLM starts with ambiguity analysis before making predictions, 
3) Few-shot w/o CoT \citep{dong2022survey}, where the LLM is evaluated by providing examples, 
4) Few-shot w/ CoT, where the LLM is evaluated by providing examples with their corresponding ambiguity analysis.
In the few-shot setting, we provide two randomly selected examples, one is ambiguous and the other is unambiguous. 
Importantly, we carefully selected 3 prompts and test all LLMs on these prompts. We present the average performance across various prompts to guarantee the statistical significance of the experimental findings.
Details on prompts are presented in Appendix \ref{four_prompt_schemes}.

% overall里面放general和小模型是hp
% 后面是深入分析为什么他们不得行

\begin{table*}[t!]
  \centering
    \scalebox{1}{
    \resizebox{\linewidth}{!}{
\begin{tabular}{ccccccccccc}
\toprule
\multirow{2}{*}{Methods} & \multicolumn{2}{c}{Zero-shot w/o CoT}            & \multicolumn{2}{c}{Zero-shot w/ CoT}             & \multicolumn{2}{c}{Few-shot w/o CoT}             & \multicolumn{2}{c}{Few-shot w/ CoT}              & \multicolumn{2}{c}{Average Performance}          \\  \cmidrule(lr){2-3} \cmidrule(lr){4-5} \cmidrule(lr){6-7}  \cmidrule(lr){8-9} \cmidrule(lr){10-11}                          & Acc.                               & F1                              & Acc.                            & F1                              & Acc.                            & F1                              & Acc.                               & F1                                 & Acc.                            & F1                              \\ \midrule
Vicuna-13B               & 50.62                              & 39.97                           & 54.75                           & 51.13                           & 50.25                           & 36.62                           & 53.50                              & \textbf{52.23}    & 52.28                           & 44.99                           \\
Llama2-13B-I              & 45.66                              & 43.64                           & 45.57                           & 45.29                           & 47.13                           & 47.04                           & 45.97                              & 42.26                              & 46.08                           & 44.56                           \\
Llama2-13B                & \textbf{55.47}    & 50.99                           & 50.97                           & 36.80                           & 46.56                           & 35.08                           & 52.19                              & 45.15                              & 51.30                           & 42.01                           \\
Llama2-70B                & 50.37                              & 34.27                           & 53.06                           & 40.29                           & 46.66                           & 39.64                           & \textbf{54.93}    & 45.42                              & 51.26                           & 39.91                           \\
ChatGPT             & \underline{54.34} & \textbf{53.45} & \textbf{57.38} & \textbf{56.91} & \textbf{51.66} & \textbf{49.28} & \underline{53.60} & \underline{51.42} & \textbf{54.25} & \textbf{52.77} \\ 

\bottomrule
\end{tabular}
}}
\caption{Overall ambiguity identification evaluation of LLMs with varying prompting schemes. ChatGPT emerges as the superior model, yet there is still considerable room for improvement, even enhanced by the CoT and Few-Shot.}
\label{main-result}
% \vspace{-2mm} 
\end{table*}

\section{Task 1: Identifying Ambiguity}
\label{task1}
This section aims to evaluate the ability of LLMs to identify different categories of ambiguous user queries, focusing on both the overall performance (cf. Section \ref{overll}) and performance specific to each category (cf. Section \ref{find}).
Following \citep{hu2023large, deng-etal-2023-prompting}, we adopt the Accuracy and F1 score as metrics. 
% Additionally, we involve two metrics, Acc@1 and Acc@0, to investigate the identification accuracy on ambiguous and unambiguous queries, respectively.

% \vspace{-1mm}
\subsection{Overall Evaluation}
\label{overll}
As shown in Table \ref{main-result}, our findings suggest that current LLMs, despite leveraging CoT and few-shot prompting, face challenges in identifying ambiguities. Our detailed observations are as follows.

\begin{figure}[ht] %H为当前位置，!htb为忽略美学标准，htbp为浮动图形
\includegraphics[width=.45\textwidth, height=0.25\textwidth]{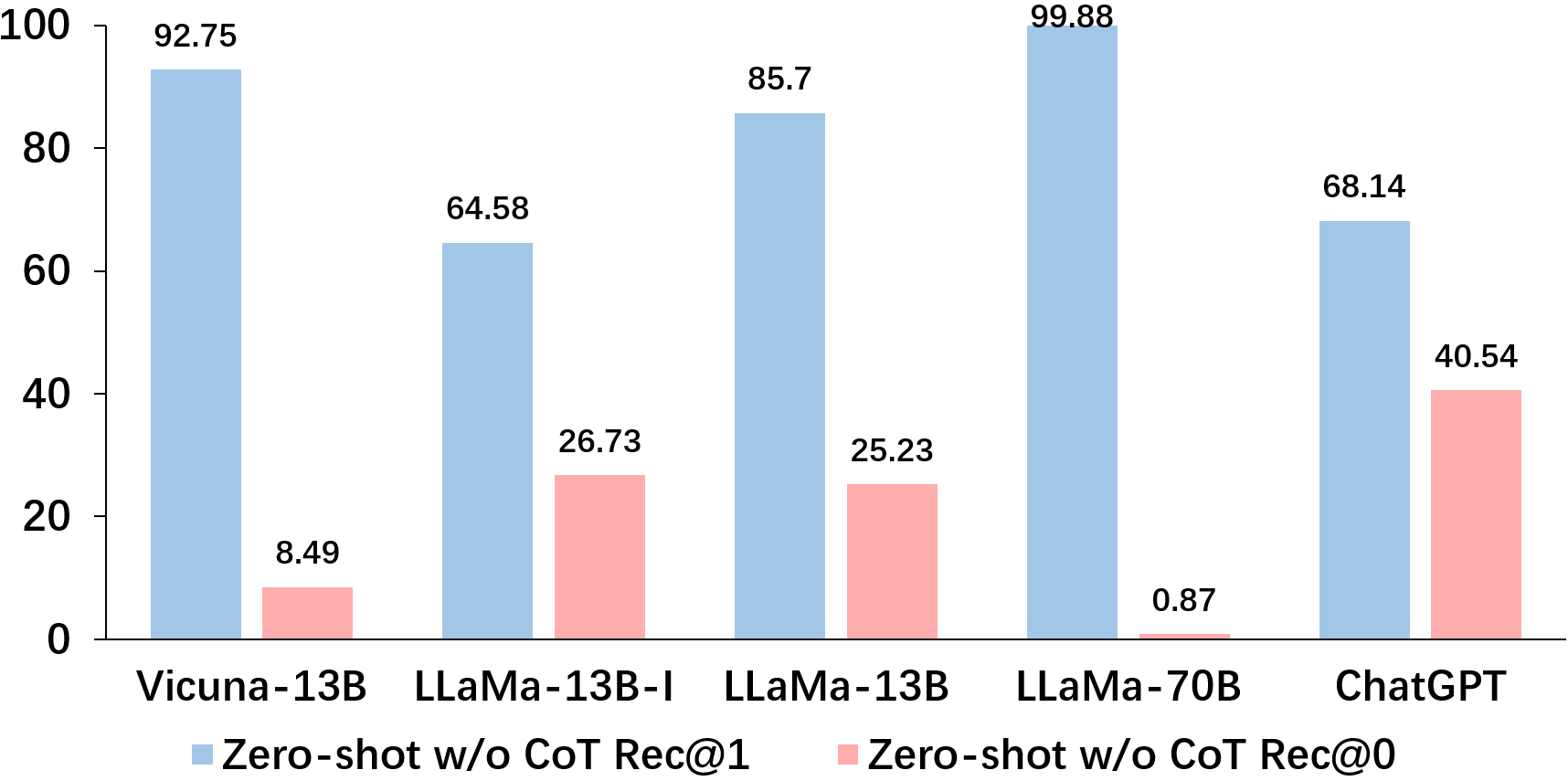} %插入图片，[]中设置图片大小，{}中是图片文件名
\centering
\caption{Investigation on the identification accuracy when handling ambiguous (i.e, Acc@1) versus unambiguous queries (i.e, Acc@0). We report the results under Zero-shot w/o CoT setting. Small-scale LLMs tend to classify most queries as ambiguous.} 
\label{Prediction_Bias} %用于文内引用的标签
% \vspace{-3mm}
\end{figure}

\noindent \textbf{In general, current LLMs are struggle to identify ambiguities}.
We observe that small-scale LLMs are unable to differentiate between ambiguous and unambiguous queries. 
In particular, they not only show significantly low performance but also demonstrate a substantial discrepancy between accuracy and F1 score. 
For instance, the accuracy of Llama2-70B with Zero-shot w/o CoT is 50.37, while its F1 score is notably lower at 34.27. 
This implies a notable variation in their performance when handling ambiguous versus unambiguous queries. 
As depicted in Figure \ref{Prediction_Bias}, these models tend to classify most queries as ambiguous, even those that are actually unambiguous. 
Compared to small-scale LLMs, ChatGPT stands out as the superior model. However, it only reaches an accuracy of 54.25\% and an F1 score of 52.77\%. There remains large room for improvement.

\begin{table}[t]
\resizebox{\linewidth}{!}{
\begin{tabular}{cccc|c}
\toprule
Metric        & Model & Zero-shot w/o CoT & Zero-shot w/ CoT & Difference \\ \hline
\multirow{5}{*}{ECE $\downarrow$} & Vicuna-13B                & 21.47             & 19.81            & -1.66      \\
                                  & Llama2-13B-I               & 22.43             & 19.91            & -2.52      \\
                                  & Llama2-13B                 & 28.48             & 45.14            & \greyhighlight{+16.66}     \\
                                  & Llama2-70B                 & 48.21             & 47.24            & -0.97      \\
                                  & ChatGPT              & 29.74             & 16.30            & -13.44      \\ \hline
\multirow{5}{*}{ROC $\uparrow$}   & Vicuna-13B                & 49.73             & 51.37            & +1.64      \\
                                  & Llama2-13B-I               & 56.18             & 56.40            & +0.22      \\
                                  & Llama2-13B                 & 57.00             & 48.22            & \greyhighlight{-8.78}      \\
                                  & Llama2-70B                 & 50.74             & 56.33            & +5.59      \\
                                  & ChatGPT              & 54.35             & 57.35            & +3.00      \\ 
\bottomrule
\end{tabular}}
\caption{Overconfidence evaluation on LLMs with and without CoT. Significant differences are marked in \greyhighlight{grey}.}
\label{uncertainty_CoT}
\end{table}

\begin{table}[t]
\resizebox{\linewidth}{!}{
\begin{tabular}{cccc|c}
\toprule
Metric                            & Model        & Zero-shot w/o CoT & Few-shot w/o CoT & Difference \\ \hline
\multirow{5}{*}{ECE $\downarrow$} & Vicuna-13B   & 21.47             & 25.66            & \greyhighlight{+4.19}      \\
                                  & Llama2-13B-I  & 22.43             & 20.99            & -1.44      \\
                                  & Llama2-13B    & 28.48             & 44.10            & \greyhighlight{+15.62}     \\
                                  & Llama2-70B    & 48.21             & 31.68            & -16.53     \\
                                  & ChatGPT & 29.74             & 13.40            & -16.34     \\ \hline
\multirow{5}{*}{ROC $\uparrow$}   & Vicuna-13B   & 49.73             & 48.70            & -1.03      \\
                                  & Llama2-13B-I  & 56.18             & 56.56            & +0.38      \\
                                  & Llama2-13B    & 57.00             & 50.55            & \greyhighlight{-6.45}      \\
                                  & Llama2-70B    & 50.74             & 43.84            & \greyhighlight{-6.9}       \\
                                  & ChatGPT & 54.35             & 51.57            & -2.78      \\ 
\bottomrule
\end{tabular}}
\caption{Overconfidence evaluation on LLMs with and without few-shot prompting. Significant differences are marked in \greyhighlight{grey}.}
\label{uncertainty_few-shot}
\end{table}

\begin{figure*}[h] %H为当前位置，!htb为忽略美学标准，htbp为浮动图形
\includegraphics[width=\textwidth, height=0.3\textwidth]{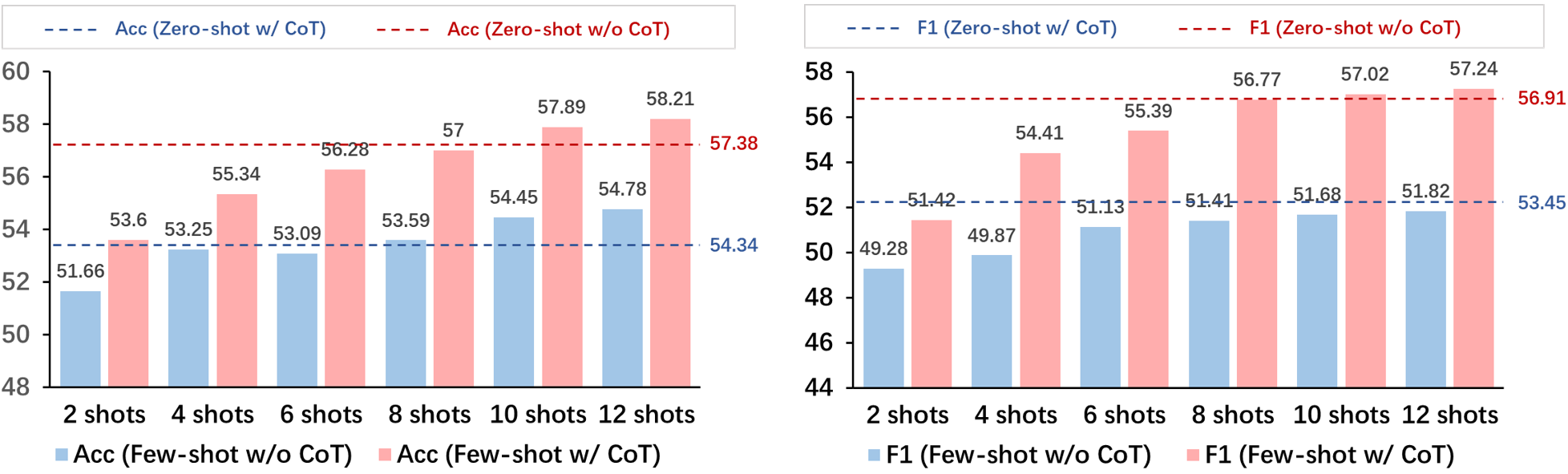} %插入图片，[]中设置图片大小，{}中是图片文件名
\centering
\caption{Performance of ChatGPT enhanced with multiple examples. We ensure a variety of categories in the examples and maintain an equal balance of ambiguous and unambiguous instances.} %最终文档中希望显示的图片标题
\label{Few_shot_analysis} %用于文内引用的标签
% \vspace{-2mm}
\end{figure*}

\begin{table*}[t!]
  \centering
    \scalebox{1}{
    \resizebox{\linewidth}{!}{
\begin{tabular}{ccccccccccccccccc}
\toprule
\multirow{3}{*}{Methods} & \multicolumn{4}{c}{Epistemic   Misalignment}     & \multicolumn{4}{c}{Linguistic   Ambiguity}                      & \multicolumn{8}{c}{Aleatoric   Output}                                  \\  \cmidrule(lr){2-5} \cmidrule(lr){6-9} \cmidrule(lr){10-17}   
                         & \multicolumn{2}{c}{contradiction} & \multicolumn{2}{c}{unfamiliar} & \multicolumn{2}{c}{lexical} & \multicolumn{2}{c}{semantic} & \multicolumn{2}{c}{what} & \multicolumn{2}{c}{whom} & \multicolumn{2}{c}{when} & \multicolumn{2}{c}{where} \\ \cmidrule(lr){2-3} \cmidrule(lr){4-5} \cmidrule(lr){6-7}   \cmidrule(lr){8-9}  \cmidrule(lr){10-11} \cmidrule(lr){12-13} \cmidrule(lr){14-15} \cmidrule(lr){16-17}
                         & Acc.       & F1         & Acc.       & F1        & Acc.          & F1           & Acc.            & F1             & Acc.        & F1         & Acc.        & F1         & Acc.        & F1         & Acc.        & F1          \\ \hline
Vicuna-13B               & 51.75      & 37.11      & 59.50      & 59.33     & \textbf{72.00}         & \textbf{71.52}        & 49.75           & 33.22          & 44.81       & 41.74      & 46.95       & 44.57      & 44.86       & 41.82      & 42.96       & 39.24       \\
Llama2-13B-I              & 49.50      & 33.11      & 46.75      & 46.47     & 52.50         & 49.20        & 48.50           & 41.31          & 30.24       & 30.14      & 31.37       & 31.32      & 27.97       & 27.72      & 29.57       & 29.44       \\
Llama2-13B                & 50.25      & 33.89      & 54.25      & 46.65     & 56.75         & 49.11        & 50.00           & 33.33          & 34.73       & 34.64      & 36.86       & 36.85      & 34.27       & 34.16      & 34.17       & 34.05       \\
Llama2-70B                & \textbf{63.25}      & \textbf{58.83}      & 50.75      & 35.81     & 55.25         & 44.04        & 50.00           & 33.33          & 31.04       & 30.77      & 31.37       & 31.07      & 31.37       & 31.07      & 31.47       & 31.16       \\
ChatGPT             & 38.00      & 28.17          & \textbf{60.00}          & \textbf{59.67}         & \underline{58.75}             & \underline{58.06}            & \textbf{50.75}               & \textbf{49.32}              & \textbf{65.40}           & \textbf{50.54}          & \textbf{68.77}           & \textbf{57.48}          & \textbf{65.00}           & \textbf{45.66}          & \textbf{63.10}           & \textbf{45.24}           \\ 
\bottomrule
\end{tabular}
}}
\caption{The fine-grained ambiguity identification evaluation results under Few-shot w/o CoT setting. ChatGPT demonstrates excellent performance across all categories of Aleatoric Output, but it does not effectively address the \textit{semantic} and \textit{contradiction} categories.}
\label{subcategory-result}
% \vspace{-6mm} 
\end{table*}

\noindent \textbf{CoT and few-shot prompting hold promise for enhancing ambiguity identification, but their effectiveness is not guaranteed.} They may lead to the overconfidence issue in small-scale LLMs, leading to negative outcomes.
As shown in Table \ref{main-result}, the effectiveness of CoT and few-shot prompting doesn't consistently improve. 
To delve deeper, we follow \citet{cole-etal-2023-selectively} and gauged LLMs' prediction confidence \footnote{Self-consistency confidence with 4 candidate answers are used to obtain the LLM's uncertainty \citep{xiong2023can}.} using Expected Calibration Error (ECE) and Area Under the Receiver Operating Characteristic curve (AUROC). ECE assesses the alignment of confidence scores with actual accuracy, while AUROC measures the ability of confidence scores to distinguish between correct and incorrect predictions. 
Our in-depth analysis, presented in Table \ref{uncertainty_CoT} and Table \ref{uncertainty_few-shot}, reveals that employing CoT and few-shot prompting leads small-scale LLMs (e.g., Llama2-13B) to exhibit overconfidence and less accurate ambiguity prediction, contrary to our intended outcome.

\noindent \textbf{Even bolstered by numerous shots and CoT support, LLMs still struggles to accurately identify query ambiguity.}
Figure \ref{Few_shot_analysis} illustrates the performance of ChatGPT when enhanced with multiple shots. 
The results indicate that the improvement seen with few-shot prompting is minimal and often inferior to the zero-shot counterpart. 
A considerable number of shots (e.g., 12 shots) are required for few-shot prompting to outperform the zero-shot method.
However, this also entails longer input lengths, risking exceeding the length limit for most small-scale LLMs in our study. 
% Merely increasing the number of shots may not enhance the ability to identify ambiguity. 
% Providing examples alone to ChatGPT could lead to the learning of superficial patterns that contradict its internal knowledge, resulting in decreased performance. Additionally, ChatGPT's difficulty in fully grasping correct reasoning with limited examples could be another contributing factor.
Providing examples alone to ChatGPT could result in the learning of superficial patterns that contradict its inherent knowledge, thereby diminishing its performance. Furthermore, ChatGPT's difficulty in fully grasping correct reasoning with limited examples could be another contributing factor.

% \noindent \textbf{It's challenging to fully comprehend the correct ambiguity reasoning with fewer examples.}
% Under the Few-shot w/o CoT setting, ChatGPT's performance decreases significantly compared to the Zero-shot w/o CoT setting, with an average 5\% decrease in accuracy and 8\% decrease in F1 score.
%We find that ChatGPT performs not well under both the Few-shot w/o CoT and Few-shot w/ CoT settings 
%To investigate this, we conduct an additional experiment by increasing the number of shots.
%As depicted in Figure \ref{Few_shot_analysis}, we reveal that ChatGPT's performance marginally improves with more examples under the Few-shot w/o CoT setting, and still remains inferior than the Zero-shot w/o CoT setting even with 12 examples (i.e., 51.82 vs. 53.45).
%We posit that providing only examples to ChatGPT could result in the teaching of superficial patterns that contradict its internal knowledge, ultimately leading to a decrease in performance.
%Furthermore, under the Few-shot w/ CoT setting, we find that there is minimal incremental improvement in performance when using a limited number of examples. 
%However, beyond a specific threshold (i.e., 10 examples), adding more examples results in better performance than the Zero-shot w/ CoT setting (i.e., 57.89 vs. 57.38 on Acc and 57.02 vs. 56.91 on F1).
% We hypothesis this phenomenon may attributed to ChatGPT's inability to fully comprehend the correct reasoning with fewer examples.

\subsection{Fine-Grained Evaluation}
\label{find}
% \vspace{-1mm}
This section analyzes the challenges LLMs faced in comprehending different ambiguities, offering insights to guide future enhancements.
Table \ref{subcategory-result} details the ambiguity identification performance of LLMs on each category. Here, we consider the Few-shot w/ CoT setting and leave more details in Appendix \ref{More_task_results}. Our observations are as follows.

\noindent \textbf{ChatGPT displays superior performance on Aleatoric Output compared to small-scale LLMs.}
Across all categories of Aleatoric Output, ChatGPT attains an average increase of 5\% in accuracy and 8\% in F1 score. 
This superior performance may stem from its vast world knowledge, enabling it to infer the absence of task-oriented elements in user queries. 
Additional results reveal ChatGPT performs exceptionally well in the "whom", while struggles more with the "when" and "where" categories. 
This suggests room for future improvement in handling queries lacking temporal and spatial elements.

\begin{table*}[t!]
  \centering
    \scalebox{1}{
    \resizebox{\linewidth}{!}{
\begin{tabular}{ccccccccccc}
\toprule
\multirow{2}{*}{Methods} & \multicolumn{2}{c}{Zero-shot w/o CoT}            & \multicolumn{2}{c}{Zero-shot w/ CoT}             & \multicolumn{2}{c}{Few-shot w/o CoT}             & \multicolumn{2}{c}{Few-shot w/ CoT}              & \multicolumn{2}{c}{Average Performance}          \\  \cmidrule(lr){2-3} \cmidrule(lr){4-5} \cmidrule(lr){6-7}  \cmidrule(lr){8-9} \cmidrule(lr){10-11}                                                   & BS                & Help.             & BS                & Help.            & BS                & Help.            & BS               & Help.            & BS                 & Help.              \\ \hline
Vicuna-13B               & 21.63             & 36.64             & 24.42             & 40.97            & 20.45             & 33.97            & 17.65            & 31.87            & 21.04              & 35.86              \\
Llama2-13B-I              & 19.65             & 33.07             & 14.29             & 26.34            & 8.01              & 14.16            & 7.96             & 17.36            & 12.48              & 22.73              \\
Llama2-13B                & 21.46             & 35.38             & 23.46             & 40.40            & 10.79             & 19.05            & 23.68            & 40.75            & 19.85              & 33.89              \\
Llama2-70B                & 22.05             & 37.35             & 19.17             & 32.72            & 19.71             & 32.80            & 22.49            & 39.99            & 20.86              & 35.71              \\
ChatGPT                  & \textbf{27.47}    & \textbf{46.45}    & \textbf{30.22}    & \textbf{50.47}   & \textbf{31.16}    & \textbf{51.58}   & \textbf{33.48}   & \textbf{53.29}   & \textbf{31.33}     & \textbf{50.45}     \\

\bottomrule
\end{tabular}
}}
\caption{Overall ambiguity clarification evaluation of LLMs with varying prompting schemes. ChatGPT emerges as the superior model to other open-sourced LLMs. We report BertScore (i.e., BS) and Help.}
\label{cq-main-result}
% \vspace{-5mm} 
\end{table*}

\noindent \textbf{The \textit{semantic} category presents a significant challenge for all LLMs.}
As shown in Table \ref{subcategory-result}, all LLMs exhibit subpar performance when dealing with ambiguous queries requiring semantic comprehension.
This indicates that current LLMs struggle to use contextual cues to clarify pronouns, highlighting their inadequacy in robustly understanding context and inferring underlying ambiguity.

\noindent \textbf{ChatGPT lags behind other small-scale LLMs on the \textit{contradiction} category.}
As shown in Table \ref{subcategory-result}, ChatGPT only achieves limited accuracy (i.e., 38) and a low F1 score (i.e., 28.17). We observe that 81.97\% of errors are false negatives, indicating that ChatGPT often misidentifies queries with self-contradictions as unambiguous. This limitation could be attributed to its training approach (i.e., SFT and RLHF \citep{ouyang2022training}), which compels ChatGPT to generate responses for all user queries, irrespective of potential contradictions.

\section{Task 2: Asking Clarifying Questions}
% \vspace{-1mm}
\label{task2}
This section investigates the ability of LLMs to produce effective clarifying questions for resolving ambiguities. 
Overall, current LLMs fail to ask high-quality clarifying questions, due to the inability of assessing their knowledge boundaries. 
Detailed observations are outlined below.

% \vspace{-2mm}
\subsection{Overall Evaluation}
\label{overll2}
We utilize \textit{BertScore} for automated assessment, as lexical matching metrics can not adequately capture clarification abilities \citep{guo2021abg}. 
Specifically, we compute the semantic similarity using BERT between the generated question and annotated clarifying questions.
Additionally, we also conduct human evaluation\footnote{Refer to Appendix \ref{Human_Evaluation} for details on human evaluation.} to score whether the generated question is helpful in resolving query ambiguity (denoted as \textbf{Help.}\footnote{It entails assigning a binary score (0 or 1) to each generated question.})

% conduct human evaluation\footnote{Refer to Appendix \ref{Human_Evaluation} for details on human evaluation.} to assess the usefulness of the generated questions in resolving ambiguities  (denoted as \textit{Help}\footnote{It involves assigning a binary score (0 or 1) to each generated clarifying question to evaluate its usefulness in resolving query ambiguity.}).

\noindent \textbf{ChatGPT demonstrating its superior capabilities in generating clarifying questions compared to small-scale LLMs.}
Table \ref{cq-main-result} showcases the effectiveness of clarifying questions produced by different LLMs. It is evident that ChatGPT demonstrates an average performance improvement of 10.29 compared to Vicuna-13B, the top-performing small-scale LLM. This indicates that ChatGPT excels in generating natural and useful clarifying questions (i.e., what to ask).

% \vspace{-1mm}
\subsection{Fine-Grained Evaluation}
\label{find2}
We provide an in-depth error analysis to reveal the inadequacies in asking clarifying questions. Since \textit{ChatGPT + Few-shot w/ CoT} stands as the most effective model, our analysis focus on it. Specifically, we randomly sampled 50 error clarifying questions (whose \textit{Help} scores are 0) from each category, 400 in total.
Inspired by \citep{deng-etal-2023-prompting}, we categorize these failure cases into four groups:
\vspace{-\topsep}
\begin{itemize}[leftmargin=*, itemindent=0.05cm, itemsep=-4pt]
    \item \textit{Wrong Aspect}. It refers the case when the generated question is aimed to clarify an incorrect aspect of the user's query.
    \item \textit{Under-specified}. The generated question is too unspecific, making it difficult for the user to provide useful feedback.
    \item \textit{Over-specified}. The generated question is an overly detailed one when the needed information is already evident in the user's original query.
    \item \textit{Generation error}. ChatGPT doesn't generate the output as the required format, such as no clarification question.
\end{itemize}
% Furthermore, we observe that the quality of clarifying questions produced by ChatGPT remain subpar, with the Help. of just 53.29 under the Few-shot w/ CoT setting.
% To find out the reason why ChatGPT fall short of generating clarifying questions, we randomly sampled 50 error cases from each category (all cases are generated by ChatGPT under the Few-shot w/ CoT setting).

\begin{figure}[ht] %H为当前位置，!htb为忽略美学标准，htbp为浮动图形
\includegraphics[width=.46\textwidth, height=0.4\textwidth]{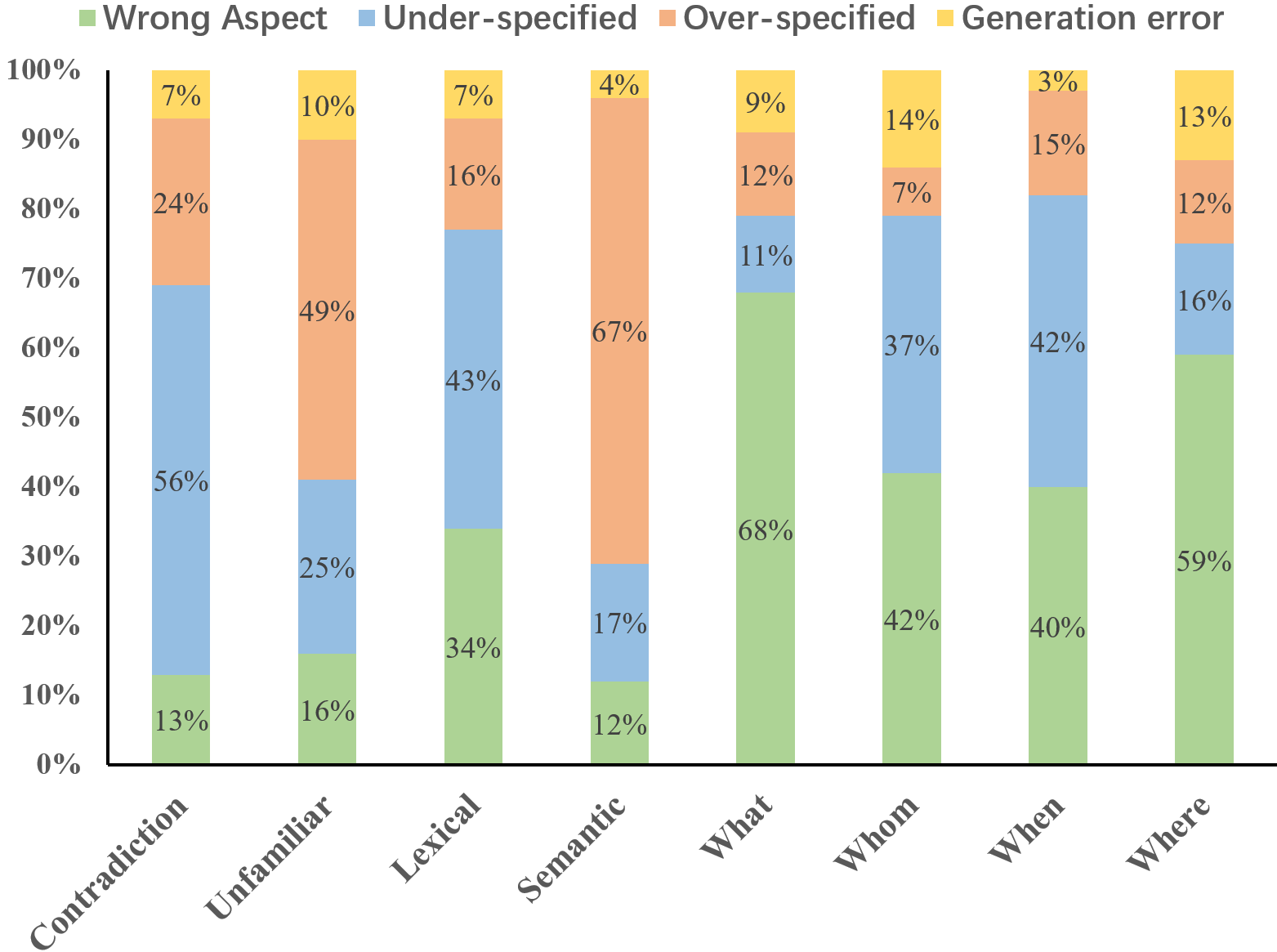} %插入图片，[]中设置图片大小，{}中是图片文件名
\centering
\caption{The statistics of error analysis. ChatGPT is unable to recognize their knowledge gap for the inadequacies in asking the effective clarifying questions.} %最终文档中希望显示的图片标题
\label{cq_error_analysis} %用于文内引用的标签
% \vspace{-6mm}
\end{figure}

As illustrated in Figure \ref{cq_error_analysis}, \textbf{inability of knowing their knowledge gap is the main reason for the inadequacies in asking effective clarifying questions}. Specifically, when dealing with the Epistemic Misalignment and Linguistic Ambiguity, most errors are concentrated on \textit{Under-specified} and \textit{Over-specified}, while \textit{Wrong Aspect} is evident in Aleatoric Output, with an average of 52.25\% error rate.
This indicates that ChatGPT can not fully comprehend semantic nuances and lack of conflict resolution despite their large parameters.
Moreover, ChatGPT use their inherent knowledge inaccurately to clarify the missing elements of ambiguous queries.
These findings imply that there exists a gap between inherent knowledge within LLMs and the ambiguities contained in user queries.

% This indicates that they can not fully comprehend semantic nuances and use context to resolve conflicts and despite their large parameters. 
% Additionally, they can not apply their vast world knowledge to accurately infer the missing elements of ambiguous queries.
% Overall, though broad world knowledge aids LLMs, our study highlights opportunities to enhance semantic mastery and reasoning abilities to precisely identify and leverage the necessary knowledge. Refining these areas could significantly advance large language models in the future.

% \subsubsection{Case Study}

\section{Conclusion}
% \vspace{-2mm}
In this work, we introduce CLAMBER, a benchmark for evaluating LLMs in identifying and clarifying ambiguous user queries through a well-organized taxonomy. 
CLAMBER comprises $\sim 12K$ high-quality data covering a wide range of ambiguity categories.
With CLAMBER, we assess strengths, weaknesses, and potential risks of various off-the-shelf LLMs. 
Our results indicate that current LLMs still face difficulties in achieving optimal performance in ambiguity identification and clarification, limiting their practical utility in advanced information search applications.
In this paper, CLAMBER acts as a foundation for enhancing the proactive capabilities of LLMs in addressing ambiguity. 
Moving forward, we plan to integrate more challenging and comprehensive datasets into our CLAMBER based on our taxonomy.

\section*{Acknowledgements}
This work was supported in part by the National Natural Science Foundation of China (No. 62272330 and No. 62206191);
in part by the Natural Science Foundation of Sichuan (No. 2023NSFSC0473), and in part by the Fundamental Research Funds for the Central Universities (No. 2023SCU12089 and No. YJ202219).

\section*{Limitations}
In this section, we discuss the limitations of this work from the following perspectives:

\noindent \textbf{Sensitivity of Prompts.} 
Similar to other studies on prompting LLMs \citep{amayuelas2023knowledge, deng-etal-2023-prompting}, 
the evaluation results are likely to be sensitive to the prompts. 
While we employ three different prompts and report the average results, it is challenging to assert that they are the most suitable ones for our specific issue.
Indeed, the sensitivity of prompts and their optimality present significant research areas within LLMs, warranting further exploration in future studies.

\noindent \textbf{Limited LLMs.}
We only use 5 Large Language Models (LLMs) in our CLAMBER benchmark due to computational constraints. If given additional resources and an improved experimental environment, it would be advantageous to evaluate the performance of other LLMs, such as PaLM540B, etc., in our CLAMBER benchmark.

\bibliography{anthology,custom}
% \bibliographystyle{acl_natbib}

% \clearpage

\appendix

\section{Prompt Design}
\label{four_prompt_schemes}
Table \ref{tab:tg_example} presents our four prompting schemes used for evaluation.
In the case of few-shot prompting, we randomly choose two examples from our CLAMBER benchmark.
The demonstration of chain-of-thoughts is written by human annotators, which represents their own ambiguity analysis.

\section{Implementation Details}

% 这里改一下超链接
The implementation of our LLMs is based on Pytorch and Transformers toolkit.
In particular, for Llama2-13B-Chat\footnote{https://huggingface.co/meta-llama/Llama-2-13b-chat-hf} and Llama2-70B-Chat\footnote{https://huggingface.co/meta-llama/Llama-2-70b-chat-hf}, we adopt the official version in Huggingface.
For Llama2-13B-instruct, We adopt the version\footnote{https://huggingface.co/Expert68/Llama2\_13b\_instructed\_\par version2} that is fine-tuned on multiple instruction-following datasets.
For Vicuna-13B, we choose the Vicuna-13B-delta-v1.5 version\footnote{https://huggingface.co/lmsys/vicuna-13b-v1.5}.
% 这里改一下解码用的参数
In particular, we set the temperature to 0 for ChatGPT and 0.5 for other open-sourced LLMs.
In addition, we set the maximum number of new tokens to 128.
During inference, the decoding strategy of open-sourced LLMs is top-p sampling with a top-p of 0.8.
For the F1 score, we use the weighted F1 score as our metric, given the balanced nature of our test set.
Our aim is to ensure the model's accuracy without ambiguity, minimizing the need for excessive clarification.
All of our experiments are conducted on two NVIDIA A100 GPUs.

\section{Details of Data Collection}
\label{detail_data_collection}
In this section, we describe the detailed data collection process of each category, including the data processing and the prompts used by GPT-4.

\subsection{ALCUNA Dataset}
The ALCUNA dataset \citep{yin2023alcuna} creates new entities by altering
existing entity attributes and relationships, resulting in artificial entities that are distinct from real-world entities.
It contains numerous question-answer pairs designed as a benchmark to evaluate the capabilities of LLMs, especially in handling new knowledge.
Specifically, we classify questions containing new entities in this dataset as ambiguous queries, and those involving existing entities as unambiguous queries.
Furthermore, we randomly select 1500 ambiguous queries and employ GPT-4 to generate a clarifying question for each one, focusing on the ambiguity of the new entity.
We provide the data examples and the prompt of generating clarifying question in Table \ref{tab:alcuna_prompts}.

\subsection{AmbiTask Dataset}
The AmbiTask Dataset \citep{tamkin2022task} constructs multiple classification tasks, each accompanied by an instruction and two provided examples. 
The two examples can lead to multiple explanations of the instruction, resulting in contradictions.
In particular, we select three classification tasks: "propn negation", "religious pronoun" and "subject location", totaling 1200 instances.
we rephrase the ambiguous tasks using rule-based templates to enhance their clarity. 
These rephrased ambiguous tasks serve as our ambiguous queries.
We generate clarifying questions based on the rule-based templates for these ambiguous queries.
Additionally, we create unambiguous queries by manually modifying the instruction and make sure the two examples can lead to just one interpretation.
We also rephrase these unambiguous queries to make them clear.
The rule-based templates and data examples are provided in Table \ref{tab:ambitask_prompts}.

\subsection{AmbER Dataset}
The AmbER dataset \citep{chen-etal-2021-evaluating} includes instances of entity ambiguity, where a single name can refer to multiple entities. 
Each ambiguous entity is annotated with its different meanings, and each meaning is associated with a factual question.
Specifically, we have chosen the top-500 most frequent entities in the \textit{non-human} category from AmbER as our data source. 
We feed the ambiguous entity and its questions related to each meaning into GPT-4, which then generates ambiguous queries along with corresponding clarifying questions.
By providing these generated ambiguous queries and corresponding clarifying questions, we guide GPT-4 to produce a clear and unambiguous version. 
Further information about the prompts and data samples can be found in Table \ref{tab:amber_prompts}.

\subsection{AmbiPun Dataset}
The AmbiPun dataset \citep{mittal2022ambipun} comprises pun words that carry diverse meanings depending on the context. 
Each pun words is annotated with its various meanings.
We randomly select 500 instances as our data resource, following the same data collection process as the AmbER dataset. Please refer to Table \ref{tab:ambipun_prompts} for the prompts and data examples.

\subsection{AmbiCoref Dataset}
The AmbiCoref dataset \cite{yuan2023ambicoref} consists of minimal pairs featuring ambiguous and unambiguous referents.
This dataset extends the scope of psycholinguistic research on how individuals perceive ambiguity in specific verb structures and their arguments.
We incorporate the ambiguous and unambiguous referent of this dataset as corresponding queries into our benchmark. For those ambiguous queries, we use a template to generate a clarifying question. The templates and examples are in Table \ref{tab:ambicoref_prompts}.

\subsection{AmbigQA Dataset}
The AmbigQA dataset \citep{min-etal-2020-ambigqa} consists of ambiguous factoid questions sourced from Natural Questions \cite{kwiatkowski2019natural}.
We classify the questions with multiple answers as ambiguous while those those with a single answer are considered unambiguous.
Furthermore, we rely on the clarifying question annotations in \citep{lee2023asking}, we use the key word in their annotations and further categorize each ambiguous question manually into four categories.
We adopt their annotated clarifying questions directly.
The data examples are presented in Table \ref{tab:ambigqa_prompts}.

\subsection{Dolly Dataset}
The Dolly dataset \citep{DatabricksBlog2023DollyV2} is commonly used for instructional fine-tuning purposes. We specifically choose the instructions from the open-qa sub-category as they align with the task of information retrieval. Our approach involves instructing GPT-4 to differentiate between ambiguous and unambiguous queries, generating clarifying questions for the ambiguous ones, and then classifying them into our predefined categories. Please refer to Table \ref{tab:dolly_prompts} for examples of the prompts and data.

% \section{Ambiguity Datasets}
% In this section, we first undertook a comprehensive survey of numerous existing ambiguity datasets in information retrieval tasks, the summary of which is presented in Table \ref{}. 
% We find that the ambiguities tackled by these datasets generally align with the eight categories outlined in our benchmark.
% These categories are found either individually or in combination across the datasets reviewed, underscoring their relevance and significance in the domain of ambiguity resolution.
% This realization encourages us to purposefully structure our evaluation framework around these eight specific categories, ensuring that our assessment is not only comprehensive but also directly targets the fundamental difficulties encountered in current ambiguity resolution.

\begin{table*}[ht]
\centering
\resizebox{\linewidth}{!}{
\begin{tabular}{l|l|l}
\hline
\textbf{Category} & \textbf{Query}                                                                                                                                                                                 & \textbf{Clarifying Question}                                                  \\ \hline
UNFAMILIAR        & Find the price of Samsung Chromecast.                                                                                                                                                          & Do you mean Google Chromecast or Samsung Chromebook?                          \\ \hline
CONTRADICTION     & \begin{tabular}[c]{@{}l@{}}Output ’X’ if the sentence contains {[}category withhold{]} and ’Y’ otherwise. \\ The critic is in the restaurant.>X. \\ The butterfly is in the river.>Y. \\ The boar is in the theatre.>?\end{tabular} & Is the category either human or indoor location?                             
\\ \hline
LEXICAL           & Tell me about the source of Nile.                                                                                                                                                              & Are you referring to the Nile river or the board game?                        \\ \hline
SEMANTIC          & When did he land on the moon?                                                                                                                                                                  & Who do you mean by "he" in the question?                                      \\ \hline
WHO               & Suggest me some gifts for my mother.                                                                                                                                                           & What are your mother's interests, hobbies, or something she has been wanting? \\ \hline
WHEN              & How many goals did Argentina score in the World Cup?                                                                                                                                           & Which World Cup are you referring to?                                         \\ \hline
WHERE             & Tell me how to reach New York.                                                                                                                                                                 & Please tell me where your departure place is?                                 \\ \hline
WHAT              & Real name of gwen stacy in spiderman?                                                                                                                                                          & Which one are you referring to: the actress,or the character?                 \\ \hline
\end{tabular}
}
\caption{The example clarifying questions associated with ambiguous queries in Table \ref{CLAMBER_Taxonomy}. There are no discerning patterns according to the ambiguity category.}
\label{tab:example_cq}
\end{table*}

\section{Human Evaluation Details}
\label{Human_Evaluation}
To evaluate the effectiveness of clarifying questions produced by LLMs, we engage 3 annotators to conduct a human evaluation.
Each annotator is tasked with evaluating each clarifying question alongside the corresponding ambiguous query and its associated category of ambiguity.
The annotators are instructed to adhere to a specific protocol for evaluating the quality of clarifying questions:
Initially, they are to verify if the clarifying questions generated by LLMs adhere to the correct format.
Subsequently, they are to determine whether the clarifying questions effectively aid in resolving ambiguity within user queries.
In cases where a clarifying question is considered unhelpful, the annotator will categorize the failure into one of four error types as detailed in \citet{deng-etal-2023-prompting}: wrong aspect, under-specified, over-specified, or generation error.
Overall, we assess totally 400 queries and measure the inter-annotator agreement.
We achieve an inter-annotator reliability of Krippen-dorff's alpha of above 0.70 for all ambiguity categories in our taxonomy.
In Table \ref{error_analysis}, we provide examples of generated clarifying questions for each error category.

% \begin{table*}[!t]
%     \centering
%     \begin{tabular}{p{0.97\textwidth}}
%     \toprule
%     \textbf{\textit{Under-specified}}     \\
%     \midrule
%     \textbf{Question}: \\
%     What is the impact of being thrown?
%     \midrule
%     \textbf{Ground Truth Clarifying Question}: \\
%     Are you referring to the physical act of being propelled through the air or the emotional state of being confused?
%     \midrule
%     \textbf{Ground Truth Clarifying Question}: \\
%     Being thrown in what context or situation?
%     \bottomrule
%     \end{tabular}
%     \caption{Example of under-specified}
%     \label{under specified}
% \end{table*}

% \begin{table*}[!t]
%     \centering
%     \begin{tabular}{p{0.97\textwidth}}
%     \toprule
%     \textbf{\textit{Over-specified}}     \\
%     \midrule
%     \textbf{Question}: \\
%     Give me the steps to purchase a new pair of shoes.

%     \midrule
%     \textbf{Ground Truth Clarifying Question}: \\
%     Where do you want to buy new shoes, online of offline?

%     \midrule
%     \textbf{Ground Truth Clarifying Question}: \\
%     Which specific steps are you referring to?

%     \bottomrule
%     \end{tabular}
%     \caption{Example of over-specified}
%     \label{over specified}
% \end{table*}

\section{More Task Results}
\label{More_task_results}
Table \ref{zero_wo_cot}, \ref{zero_shot_w_cot}, \ref{few_shot_wo_cot} present the results of all LLMs across different categories under three different settings: Zero-shot w/o CoT, Zero-shot w/ CoT, and Few-shot w/o CoT.
We discover that while the exact values vary, the overall performance and analysis conclusions remain largely consistent with Sec \ref{find}.

\begin{table*}[t!]
  \centering
    \scalebox{1}{
    \resizebox{\linewidth}{!}{
\begin{tabular}{ccccccccccccccccc}
\toprule
\multirow{3}{*}{Methods} & \multicolumn{4}{c}{Epistemic   Misalignment}     & \multicolumn{4}{c}{Linguistic   Ambiguity}                      & \multicolumn{8}{c}{Aleatoric   Output}                                  \\  \cmidrule(lr){2-5} \cmidrule(lr){6-9} \cmidrule(lr){10-17}   
                         & \multicolumn{2}{c}{contradiction} & \multicolumn{2}{c}{unfamiliar} & \multicolumn{2}{c}{lexical} & \multicolumn{2}{c}{semantic} & \multicolumn{2}{c}{what} & \multicolumn{2}{c}{whom} & \multicolumn{2}{c}{when} & \multicolumn{2}{c}{where} \\ \cmidrule(lr){2-3} \cmidrule(lr){4-5} \cmidrule(lr){6-7}   \cmidrule(lr){8-9}  \cmidrule(lr){10-11} \cmidrule(lr){12-13} \cmidrule(lr){14-15} \cmidrule(lr){16-17}
                                                 & Acc.            & F1              & Acc.           & F1            & Acc.          & F1          & Acc.          & F1           & Acc.        & F1         & Acc.        & F1         & Acc.        & F1         & Acc.        & F1          \\ \hline
Vicuna-13B               & 54.25           & 54.24           & 49.75          & 37.17         & 49.25         & 33.43       & 50.25         & 33.89        & 21.26       & 18.75      & 21.38       & 18.80      & 21.58       & 18.95      & 21.78       & 19.09       \\
Llama2-13B-I              & 46.00           & 32.67           & 44.75          & 43.76         & 45.50         & 44.49       & 49.75         & \textbf{48.55}        & 37.82       & 37.43      & 34.67       & 33.86      & 33.17       & 32.10      & 34.67       & 33.86       \\
Llama2-13B                & \textbf{64.25}           & \textbf{59.01}           & \textbf{50.75}          & \textbf{44.42}         & 47.50         & 41.92       & 48.25         & 33.38        & 44.01       & 43.25      & 42.26       & 41.22      & 45.55       & 45.01      & 44.66       & 44.00       \\
Llama2-70B                & 50.50           & 34.43           & 50.00          & 33.33         & 50.75         & 34.98       & 50.00         & 33.33        & 20.96       & 17.97      & 20.68       & 17.74      & 20.88       & 17.88      & 20.88       & 17.88       \\
ChatGPT                  & 39.50               & 30.10               & \textbf{50.75}              & 36.59             & \textbf{53.50}             & \textbf{49.23}           & \textbf{54.50}             & 44.75            & \textbf{49.70}           & \textbf{46.56}          & \textbf{49.95}           & \textbf{46.90}          & \textbf{52.44}           & \textbf{50.15}         & \textbf{49.35}           & \textbf{46.10}           \\ 

\bottomrule
\end{tabular}
}}
\caption{The fine-grained ambiguity identification evaluation results under Zero-shot w/o CoT setting.}
\label{zero_wo_cot}
\end{table*}

\begin{table*}[t!]
  \centering
    \scalebox{1}{
    \resizebox{\linewidth}{!}{
\begin{tabular}{ccccccccccccccccc}
\toprule
\multirow{3}{*}{Methods} & \multicolumn{4}{c}{Epistemic   Misalignment}     & \multicolumn{4}{c}{Linguistic   Ambiguity}                      & \multicolumn{8}{c}{Aleatoric   Output}                                  \\  \cmidrule(lr){2-5} \cmidrule(lr){6-9} \cmidrule(lr){10-17}   
                         & \multicolumn{2}{c}{contradiction} & \multicolumn{2}{c}{unfamiliar} & \multicolumn{2}{c}{lexical} & \multicolumn{2}{c}{semantic} & \multicolumn{2}{c}{what} & \multicolumn{2}{c}{whom} & \multicolumn{2}{c}{when} & \multicolumn{2}{c}{where} \\ \cmidrule(lr){2-3} \cmidrule(lr){4-5} \cmidrule(lr){6-7}   \cmidrule(lr){8-9}  \cmidrule(lr){10-11} \cmidrule(lr){12-13} \cmidrule(lr){14-15} \cmidrule(lr){16-17}
                        & Acc.            & F1              & Acc.           & F1            & Acc.          & F1          & Acc.          & F1           & Acc.        & F1         & Acc.        & F1         & Acc.        & F1         & Acc.        & F1          \\ \hline
Vicuna-13B               & \textbf{67.75}           & \textbf{67.41}           & 52.25          & 44.79         & 56.50         & 51.54       & 52.25         & 44.56        & 37.23       & 37.00      & 36.16       & 35.85      & 36.86       & 36.61      & 38.16       & 38.01       \\
Llama2-13B-I              & 39.50           & 28.32           & 48.50          & 48.37         & 48.50         & 48.03       & 53.75         & \textbf{53.65}        & 38.32       & 36.52      & 37.86       & 35.96      & 37.96       & 36.08      & 38.26       & 36.46       \\
Llama2-13B                & 51.25           & 36.46           & 50.50          & 34.43         & 49.50         & 33.11       & 50.00         & 33.33        & 24.45       & 22.86      & 24.18       & 22.61      & 24.08       & 22.53      & 24.88       & 23.17       \\
Llama2-70B                & 67.50           & 63.66           & 50.75          & 36.20         & 53.00         & 39.67       & 50.00         & 33.33        & 22.16       & 19.59      & 21.78       & 19.27      & 21.78       & 19.27      & 21.88       & 19.35       \\
ChatGPT                  & 42.48               & 38.97               & \textbf{55.25}              & \textbf{55.24}             & \textbf{74.00 }            & \textbf{72.79}           & \textbf{54.00}             & 43.26            & \textbf{65.70}           & \textbf{53.44}          & \textbf{64.35}           & \textbf{50.61}          & \textbf{64.00}           & \textbf{49.93}          & \textbf{63.30}           & \textbf{48.42}           \\ 

\bottomrule
\end{tabular}
}}
  \caption{The fine-grained ambiguity identification evaluation results under Zero-shot w/ CoT setting.}
\label{zero_shot_w_cot} 
\end{table*}

\begin{table*}[t!]
  \centering
    \scalebox{1}{
    \resizebox{\linewidth}{!}{
\begin{tabular}{ccccccccccccccccc}
\toprule
\multirow{3}{*}{Methods} & \multicolumn{4}{c}{Epistemic   Misalignment}     & \multicolumn{4}{c}{Linguistic   Ambiguity}                      & \multicolumn{8}{c}{Aleatoric   Output}                                  \\  \cmidrule(lr){2-5} \cmidrule(lr){6-9} \cmidrule(lr){10-17}   
                         & \multicolumn{2}{c}{contradiction} & \multicolumn{2}{c}{unfamiliar} & \multicolumn{2}{c}{lexical} & \multicolumn{2}{c}{semantic} & \multicolumn{2}{c}{what} & \multicolumn{2}{c}{whom} & \multicolumn{2}{c}{when} & \multicolumn{2}{c}{where} \\ \cmidrule(lr){2-3} \cmidrule(lr){4-5} \cmidrule(lr){6-7}   \cmidrule(lr){8-9}  \cmidrule(lr){10-11} \cmidrule(lr){12-13} \cmidrule(lr){14-15} \cmidrule(lr){16-17}
 & Acc.            & F1              & Acc.           & F1            & Acc.          & F1          & Acc.          & F1           & Acc.        & F1         & Acc.        & F1         & Acc.        & F1         & Acc.        & F1          \\ \hline
Vicuna-13B               & \textbf{50.00}           & \textbf{33.33}           & 49.50          & 38.37         & 51.50         & 38.48       & 50.50         & 35.68        & 22.46       & 20.42      & 22.58       & 20.49      & 22.58       & 20.49      & 22.18       & 20.18       \\
Llama2-13B-I              & 48.50           & 32.66           & 56.00          & 55.51         & 55.75         & 55.08       & 47.75         & 46.95        & 39.22       & 36.54      & 39.46       & 36.87      & 39.06       & 39.06      & 39.46       & 36.87       \\
Llama2-13B                & 14.75           & 12.85           & 52.50          & 41.10         & 48.25         & 33.38       & 50.00         & 33.33        & 24.45       & 22.99      & 24.58       & 23.06      & 24.78       & 23.22      & 24.88       & 23.30       \\
Llama2-70B                & \textbf{50.00}           & \textbf{33.33}           & 51.00          & 42.73         & 45.75         & 41.90       & 50.00         & 33.77        & 27.45       & 27.44      & 29.87       & 29.80      & 26.97       & 26.97      & 26.47       & 26.47       \\
ChatGPT                  & 39.00               & 28.05               & \textbf{60.00}              & \textbf{59.67 }            & \textbf{58.75}             & \textbf{58.06}           & \textbf{50.75}             & \textbf{49.32}            & \textbf{65.40}           & \textbf{50.54}          & \textbf{68.77}           & \textbf{57.48}          & \textbf{65.00}           & \textbf{49.66}          & \textbf{63.10}           & \textbf{45.24}           \\ 

\bottomrule
\end{tabular}
}}
\caption{The fine-grained ambiguity identification evaluation results under Few-shot w/o CoT setting.}
\label{few_shot_wo_cot}
\end{table*}

\section{Human Validation and Revision}
\label{Manual_Revision}
We initially engage 8 language experts via online platforms. 
Subsequently, they are assigned the task of reviewing 50 data samples according to provided instructions as part of a qualifying assessment. 
The 5 experts who successfully pass this assessment are then designated to validate and revise our dataset. 
For each query, they are given the respective ambiguity label and a corresponding clarifying question if the query is ambiguous.
They are required to adhere to a specific protocol for validating and revising our dataset: 
Firstly, they need to verify if the query is ambiguous and if the ambiguity label assigned is accurate. 
Secondly, if the query is deemed ambiguous, they should evaluate whether the clarifying question effectively resolves any ambiguity. 
In instances of differing opinions during validation, discussions should be held to reach a consensus on the final data outcome. 
If significant disagreement persists even after discussion, the data will be discarded.
We ensure the quality of our final data in two ways. 
The two authors of this paper acted as meta-reviewers, selecting 50 questions from each of the eight categories across the three dimensions in CLAMBER.
The meta-reviewers assessed the correctness of ambiguity labels and the effectiveness of clarifying questions. For the 400 data samples, the average label accuracy was 92.4\% and the average BLEU score was 73.2. 
Based on the results from the meta-reviewers, the data in CLAMBER is considered to be of high quality.

\begin{table*}[!t]
    \centering
    \begin{tabular}{p{0.97\textwidth}}
    \toprule
    \textbf{\textit{Wrong Aspect}}     \\
    \midrule
    \textbf{Question}: \\
    What is the dry mass of a seed of Dicent canadensis?\\
    \textbf{Ground Truth Clarifying Question}: \\
    What is the intended meaning of "Dicent canadensis"? Is it a specific plant species or a misspelling of a different plant name?\\
    \textbf{LLM-generated Clarifying Question}: \\
    Are you asking for the average dry mass of a seed of Dicentra canadensis or the specific dry mass of a particular seed? \\
    \midrule
    \textbf{\textit{Under-Specified}}     \\
    \midrule
    \textbf{Question}: \\
    What is the impact of being thrown?\\
    \textbf{Ground Truth Clarifying Question}: \\
    Are you referring to the physical act of being propelled through the air or the emotional state of being confused?\\
    \textbf{LLM-generated Clarifying Question}: \\
    Being thrown in what context or situation? \\
    \midrule
    \textbf{\textit{Over-Specified}} \\
    \midrule
    \textbf{Question}: \\
    Give me the steps to purchase a new pair of shoes.\\
    \textbf{Ground Truth Clarifying Question}: \\
    Where do you want to buy new shoes, online of offline?\\
    \textbf{LLM-generated Clarifying Question}: \\
    Which specific steps are you referring to? \\
    \midrule
    \textbf{\textit{Generation Error}}     \\
    \midrule
    \textbf{Question}: \\
    What are some common symptoms of someone who is moped?\\
    \textbf{Ground Truth Clarifying Question}: \\
    Are you referring to someone who is feeling apathetic, gloomy, or dazed, or are you referring to a motorbike that can be pedaled or driven by a low-powered gasoline engine?\\
    \textbf{LLM-generated Clarifying Question}: \\
    \# \\
    \bottomrule
    \end{tabular}
    \caption{Examples of Error Analysis}
    \label{error_analysis}
\end{table*}

\begin{table*}[!t]
    \centering
    \begin{tabular}{p{0.97\textwidth}}
    \toprule
    \textbf{\textit{Prompting Schemes}}     \\
    \midrule
    \textbf{Zero-shot w/o CoT}: Given a query, answer the question or ask a clarifying question. The response should start with "The answer is" or "The clarifying question is".\\
    Question: <Question> \\
    \midrule
    \textbf{Zero-shot w/ CoT}: Given a query first identify whether the question is ambiguous or not. If it is ambiguous, ask a clarifying question. If it is not ambiguous, answer the question. The response should start with the ambiguity analysis of the question and then follow by "Therefore, the question is not ambiguous. The answer is" or "Therefore, the question is ambiguous. The clarifying question is".\\
    Question: <Question> \\
    \midrule
    \textbf{Few-shot w/o CoT}: Given a query, answer the question or ask a clarifying question. The response should start with "The answer is" or "The clarifying question is".\\ \\
    Question: Who played michael myers in rob zombie's movie?\\
    Output: The answer is Tyler Mane.\\ \\
    Question: Give me some Mother's Day gift ideas.\\
    Output: The clarifying question is What are your mother's interests, hobbies, or something she has been wanting?\\ \\
    
    Question: <Question> \\
    \midrule
    \textbf{Few-shot w/ CoT}: Given a query, answer the question or ask a clarifying question. The response should start with "The answer is" or "The clarifying question is".\\ \\
    Question: Who played michael myers in rob zombie's movie?\\
    Output: In Rob Zombie's "Halloween" films, the role of Michael Myers was primarily played by Tyler Mane. Mane portrayed the iconic horror character in both Rob Zombie's "Halloween" (2007) and its sequel, "Halloween II" (2009). Therefore, the question is not ambiguous. The answer is Tyler Mane.\\ \\
    Question: Give me some Mother's Day gift ideas.\\
    Output: there could be underlying ambiguities depending on the interests of the specific mother in question, the budget, and the giver's relationship to the mother. Without this context, a broad range of ideas could be considered appropriate. Therefore, the question is ambiguous. The clarifying question is: What are the interests or hobbies of the mother, and is there a particular budget range for the gift?\\ \\
    
    Question: <Question> \\
    \bottomrule
    \end{tabular}
    \caption{Four prompting schemes for ambiguity identification and clarification.}
    \label{tab:tg_example}
\end{table*}

\begin{table*}[!t]
    \centering
    \begin{tabular}{p{0.97\textwidth}}
    \toprule
    \textbf{\textit{ALCUNA dataset}}     \\
    \midrule
    \textbf{The prompt of generating clarifying questions}: \\
    Given the user question: <question>. \\
    Note that the <entity> is a non-existent entity fabricated by existing entities.\\
    You need to generate a clarifying question about the <ENTITY> to better know its intended meaning. \\ \\
    Your Generated Clarifying Question: \\
    \midrule
    \textbf{Data Examples}: \\
    An ambiguous example \\
    Query: What is the latitude of the habitat of inyidiidae? \\
    Clarifying Question: Can you please provide more information about "inyidiidae"? \\ \\
    A unambiguous example \\
    Query: Is Mozambique a geographic distribution of Mantodea? \\
    \bottomrule
    \end{tabular}
    \caption{The prompt and data examples of the ALCUNA dataset}
    \label{tab:alcuna_prompts}
\end{table*}

\begin{table*}[!t]
    \centering
    \begin{tabular}{p{0.97\textwidth}}
    \toprule
    \textbf{\textit{AmbiCoref dataset}}     \\
    \midrule
    \textbf{The template of clarifying questions}: \\
    What does <PRONOUN> refer to? <A> or <B>? \\
    \midrule
    \textbf{Data Examples}: \\
    An ambiguous example \\
    Query: Matthew bought Joshua a pizza after he asked for more food. Who asked for more food? \\
    Clarifying Question: What does he refer to? Matthew or Joshua? \\ \\
    A unambiguous example \\
    Query: Matthew made Joshua a square pizza before he submitted the order.
Who submitted the order? \\
    \bottomrule
    \end{tabular}
    \caption{The template and data examples of the AmbiCoref dataset}
    \label{tab:ambicoref_prompts}
\end{table*}

\begin{table*}[!t]
    \centering
    \begin{tabular}{p{0.97\textwidth}}
    \toprule
    \textbf{\textit{AmbiTask dataset}}     \\
    \midrule
    \textbf{The template of rephrasing ambiguous queries}: \\
    The all possible word categories are either <category 1> or <category 2>.\\The following two examples share a specific word category. You need to first infer the specific word category from the examples. \\Please output "X" if the given sentence mentions the specific word category. Please output "Y" if the given sentence does not mention the word category.\\ \\Examples: \\The photographer is not in the restaurant.\\Bernie Sanders has been in the theatre.\\ \\The Given Sentence: \\Paul Atreides may not be in the hotel lobby. \\
    \midrule
    \textbf{The template of rephrasing unambiguous queries}: \\
    Please output "X" if the given sentence contains a word of <category>. Please output "Y" if the given sentence does not contain any word of <category>.\\ \\ Examples: \\ The fugitive has not been in the museum.(Output: X)\\ Noam Chomsky was in the film studio.(Output: Y)\\ \\ The Given Sentence: \\ The hiker was in the laboratory. \\
    \midrule
    \textbf{The template of clarifying questions}: \\
    Is the category either <category 1> or <category 2>? \\
    \midrule
    \textbf{Data Examples}: \\
    An ambiguous example \\
    Query: The all possible word categories are either "does not contain a negation" or "proper noun".\\The following two examples share a specific word category. You need to first infer the specific word category from the examples. \\Please output "X" if the given sentence mentions the specific word category. Please output "Y" if the given sentence does not mention the word category.\\ \\Examples: \\The photographer is not in the restaurant.\\Bernie Sanders has been in the theatre.\\ \\The Given Sentence: Paul Atreides may not be in the hotel lobby. \\
    Clarifying Question: Is the category either does not contain a negation or proper noun? \\ \\
    A unambiguous example \\
    Query: Please output "X" if the given sentence contains a word of "common noun". Please output "Y" if the given sentence does not contain any word of "common noun".\\ \\ Examples: \\ The fugitive has not been in the museum.(Output: X)\\ Noam Chomsky was in the film studio.(Output: Y)\\ \\ The Given Sentence: The hiker was in the laboratory. \\
    \bottomrule
    \end{tabular}
    \caption{The prompt, clarifying question template and data examples of the AmbiTask dataset}
    \label{tab:ambitask_prompts}
\end{table*}

\begin{table*}[!t]
    \centering
    \begin{tabular}{p{0.97\textwidth}}
    \toprule
    \textbf{\textit{AmbER dataset}}     \\
    \midrule
    \textbf{The prompt of generating ambiguous queries and clarifying questions}: 
    \\
    \#\#\#\#\#\#\#\#\#\#\#\#\#\#\# \\ <QUESTION 1> \\ <QUESTION 2> \\ \#\#\#\#\#\#\#\#\#\#\#\#\#\#\#  \\According to the above example questions, Note that <ENTITY> is an ambiguous entity and has multiple meanings. \\ You should generate a new question using the <ENTITY> and random context. \\ You need to make sure the generated question is ambiguous and answering the generated question requires further clarification. \\ FORMAT: \{"question": <STRING>, "clarifying\_question": <STRING>\} \\
    \midrule
    \textbf{The prompt of generating unambiguous queries}: 
    \\
     Given an ambiguous query and its clarifying question, you need to generate a unambiguous query based on them. \\ FORMAT: \{"unambiguous query": <STRING>\} \\
    \midrule
    \textbf{Data Examples}: \\
    An ambiguous example \\
    Query: What is the history of Alcatraz? \\
    Clarifying Question: Are you referring to the history of the Alcatraz Island or the history of the band Alcatraz? \\ \\
    A unambiguous example \\
    Query: What are the tracks in the album or soundtrack called Birds? \\
    \bottomrule
    \end{tabular}
    \caption{The prompt and data examples of the AmbER dataset}
    \label{tab:amber_prompts}
\end{table*}

\begin{table*}[!t]
    \centering
    \begin{tabular}{p{0.97\textwidth}}
    \toprule
    \textbf{\textit{AmbiPun dataset}}     \\
    \midrule
    \textbf{The prompt of generating ambiguous queries and clarifying questions}: 
    \\ //1. Generate ambiguous queries
    Given a polysemy word <WORD>, it has two senses, including of <SENSE1> and <SENSE2>. \\ You need to generate an information-seeking question based on the word <WORD>. \\ You need to make the generated question be ambiguous due to the polysemy of word <WORD>. \\ Note the question needs to contain the word <WORD>. \\ Answering the generated requires a clarifying question to better understand the word <WORD>. \\ generated question: \\ \\ //2. Generate clarifying question \\ Given a question: <QUESTION> \\ Note the polysemy word <WORD> has two senses, including of <SENSE1> and <SENSE2>. \\ The given question has ambiguity due to the polysemy word <WORD>. \\ You need to generate a clarifying question based on the word <WORD> to better clarify the ambiguity of the given question. \\ clarifying question: \\
    \midrule
    \textbf{The prompt of generating unambiguous queries}: 
    \\
     Given an ambiguous query and its clarifying question, you need to generate a unambiguous query based on them. \\ FORMAT: \{"unambiguous query": <STRING>\} \\
    \midrule
    \textbf{Data Examples}: \\
    An ambiguous example \\
    Query: What is the meaning of Smart? \\
    Clarifying Question: Are you referring to the adjective 'smart' or a specific brand called 'Smart'? \\ \\
    A unambiguous example \\
    Query: What are the common strategies for saving money? \\
    \bottomrule
    \end{tabular}
    \caption{The prompt and data examples of the AmbiPun dataset}
    \label{tab:ambipun_prompts}
\end{table*}

\begin{table*}[!t]
    \centering
    \begin{tabular}{p{0.97\textwidth}}
    \toprule
    \textbf{\textit{AmbigQA dataset}}     \\
    \midrule
    \textbf{Data Examples}: \\
    An ambiguous example \\
    Query: Who played kelly on the drew carey show? \\
    Clarifying Question: Which role: Kellie Newmark, Marlo Kelly, Grace Kelly, or Kelly Walker? \\ \\
    A unambiguous example \\
    Query: Where did they film ash vs evil dead? \\
    \bottomrule
    \end{tabular}
    \caption{The prompt and data examples of the AmbigQA dataset}
    \label{tab:ambigqa_prompts}
\end{table*}

\begin{table*}[!t]
    \centering
    \begin{tabular}{p{0.97\textwidth}}
    \toprule
    \textbf{\textit{Dolly dataset}}     \\
    \midrule
    \textbf{The prompt of generating clarifying questions and category classification}: 
    \\
     Give you an instruction, you first need to judge whether the instruction is ambiguous or not.\\ If you think the instruction is ambiguous and falls into one of the following ambiguous types, \\you need to output its ambiguous type and the corresponding clarifying questions to help answer the ambiguous instruction.\\ If you think the instruction is not ambiguous and does not miss any specific information, \\you need to rewrite it and make sure it falls into one of the following ambiguous types. \\ Ambiguous types:\\ 1. Missing personal information. \\For example, the instruction "Suggest me some good movies" misses the information of the user personal preference. \\ 2. Missing spatial information. \\For example, the instruction "How to reach a destination" misses the spatial information of the departure location. \\ 3. Missing temporal information. \\For example, the instruction "Make a restaurant reservation" misses the temporal information of the reservation time.\\4. Missing specific task-related information. \\For example, the instruction "convert string to int" misses the information of the programming language.\\ \\ You should output the ambiguous type, the ambiguous instruction and its corresponding clarifying questions for each instruction. \\ FORMAT: \{"ambiguous type": <STRING>, "ambiguous instruction": <STRING>, "clarifying question": <STRING>\}\\
    \midrule
    \textbf{Data Examples}: \\
    An ambiguous example \\
    Query: Give me some Mother's Day gift ideas \\
    Clarifying Question: What are your mother's interests, hobbies, or something she has been wanting? \\ \\
    A unambiguous example \\
    Query: Top scorer of uefa champions league of all time? \\
    \bottomrule
    \end{tabular}
    \caption{The prompt and data examples of the Dolly dataset}
    \label{tab:dolly_prompts}
\end{table*}

\end{document}